\documentclass{article}

\newif\ifcameraReady
\cameraReadyfalse

\PassOptionsToPackage{numbers,sort&compress}{natbib}
\usepackage[final,main]{neurips_2026}

\usepackage[utf8]{inputenc}
\usepackage[T1]{fontenc}
\usepackage{hyperref}
\usepackage{url}
\usepackage{booktabs}
\usepackage{amsfonts}
\usepackage{amsmath}
\usepackage{amssymb}
\usepackage{nicefrac}
\usepackage{microtype}
\usepackage[table]{xcolor}
\usepackage{graphicx}
\DeclareGraphicsExtensions{.pdf,.png,.jpg}
\usepackage{subcaption}
\usepackage{algorithm}
\usepackage{algorithmic}
\usepackage{multirow}
\usepackage{makecell}
\usepackage{colortbl}
\usepackage{xspace}
\usepackage[breakable]{tcolorbox}
\usepackage{fontawesome5}

\definecolor{oursblue}{RGB}{240, 240, 240}
\definecolor{darkblue}{rgb}{0, 0, 0.5}
\definecolor{clrRetired}{HTML}{D62727}
\definecolor{clrStable}{HTML}{2172B5}
\definecolor{clrActive}{HTML}{6592C0}
\definecolor{clrTrial}{HTML}{CE7336}
\definecolor{clrGray}{HTML}{9E9E9E}
\definecolor{clrLightGray}{HTML}{BDBDBD}

\hypersetup{
  pdftitle={SkillForge: Co-Evolving Skills and Agents via Dynamic Skill Lifecycles},
  pdfauthor={Yuyao Ge, Yiwei Wang, Yuchen He, Baolong Bi, Lingrui Mei, Jiayu Yao, Lizhe Chen, Shenghua Liu},
  colorlinks=true,
  citecolor=darkblue,
  linkcolor=darkblue,
  urlcolor=darkblue
}

\newcommand{\codelink}[1]{{\fontfamily{ppl}\selectfont
  \raisebox{-4.2pt}{\includegraphics[height=1.3em]{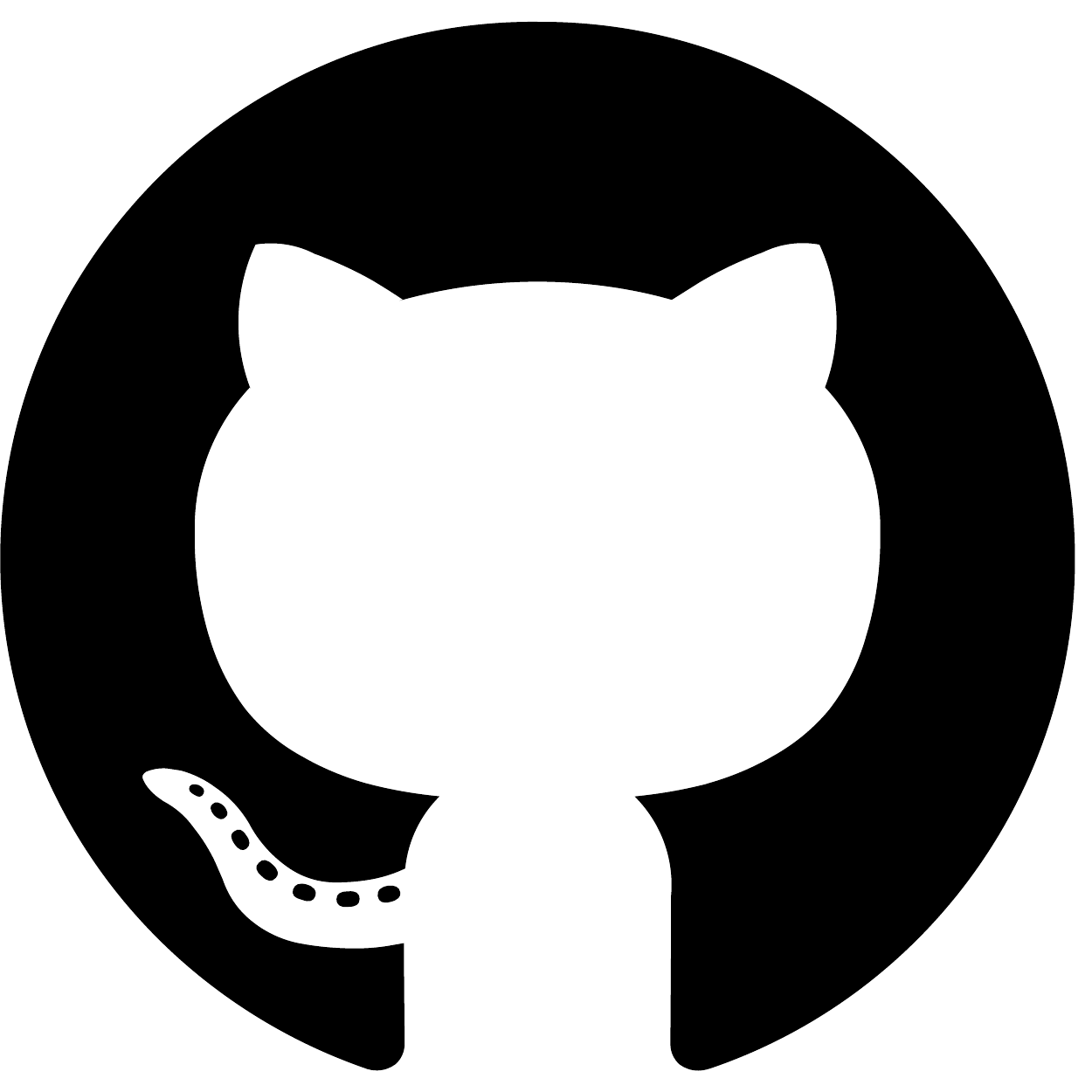}}\hspace{0.45em}{\urlstyle{same}\url{#1}}}}

\newcommand{\up}[1]{$_{\scriptscriptstyle\blacktriangle #1}$}
\newcommand{\dn}[1]{$_{\scriptscriptstyle\blacktriangledown #1}$}
\newcommand{\best}[1]{\textbf{#1}}
\newcommand{\second}[1]{\underline{#1}}
\newcommand{\ours}{\cellcolor{oursblue}}

\newcommand{\method}{\textsc{SkillForge}\xspace}
\newcommand{\skillfurnace}{\textsc{SkillFurnace}\xspace}

\title{\method: Co-Evolving Skills and Agents \\via Dynamic Skill Lifecycles}

\author{%
  Yuyao Ge$^{\clubsuit}$ \quad
  Yiwei Wang$^{\diamondsuit}$ \quad
  Yuchen He$^{\clubsuit}$ \quad
  Baolong Bi$^{\clubsuit}$ \\[3pt]
  \textbf{Lingrui Mei}$^{\clubsuit}$ \quad
  \textbf{Jiayu Yao}$^{\clubsuit}$ \quad
  \textbf{Lizhe Chen}$^{\P}$ \quad
  \textbf{Shenghua Liu}$^{\clubsuit}$\thanks{Corresponding author.} \\[6pt]
  $^{\clubsuit}$Institute of Computing Technology, Chinese Academy of Sciences \\
  $^{\diamondsuit}$University of California, Merced \quad
  $^{\P}$Tsinghua University \\[3pt]
  \texttt{\{geyuyao24z, liushenghua\}@ict.ac.cn} \\[9pt]
  \codelink{https://github.com/YuyaoGe/SkillForge}
}

\begin{document}

\maketitle

\begin{abstract}
Memory-augmented reinforcement learning strengthens LLM agents' ability to solve complex long-horizon tasks. Skills are one such form of memory, pairing instructions with an applicability condition over task types. However, retaining every skill indiscriminately as the policy improves lets obsolete or harmful entries accumulate and mislead the agent.
We propose \textbf{\method}, an agentic RL method that compiles and evolves the skill library through a fitness-driven skill lifecycle of \texttt{trial}, \texttt{active}, \texttt{stable}, and \texttt{retired} states, so that the skills and the model co-evolve throughout training. A pre-RL evaluation phase first uses the base model's own rollouts to pre-retire low-fitness skills, yielding a filtered library that then seeds supervised fine-tuning. Reinforcement learning takes over from this checkpoint, and at each iteration selective retirement, stabilization, and LLM-guided mutation continue to forge the skill library alongside policy optimization.
Across multiple interactive agent benchmarks, \method achieves the highest aggregate success rate, delivering up to 7.8\% relative improvement over the strongest baseline while keeping the skill library compact throughout training. We introduce \textbf{\skillfurnace}, a dataset of 5k+ annotated records bundling retirement-filtered SFT trajectories, evolved skill libraries with fitness annotations, and retirement events with human-annotated failure categories to support research on skill quality and lifecycle management.
\end{abstract}

\section{Introduction}
\label{sec:intro}

Generalist agents such as Claude Code~\citep{claudecode} and OpenClaw~\citep{openclaw} derive much of their capability from modular, reusable abstractions provided at inference time. Even frontier large language models (LLMs) sit behind a fixed knowledge cutoff, so they cannot absorb personalized workflows~\citep{mem0} or adapt to post-deployment tooling conventions~\citep{simplemem}. To close this gap at inference time, recent work centers on the \textit{skill}: a bundle of instructions, examples, and applicability conditions retrieved into the agent's context on demand. Skill libraries are assembled from procedural memory combining step-level and script-level abstractions~\citep{memp}, or through autonomous skill-discovery loops~\citep{exif}. As libraries grow, however, retrieval-based augmentation degrades: top-$k$ retrieval returns noisier context, outdated skills crowd out high-quality ones, and many entries no longer match the agent's current capability. \citet{li2026skillsbenchbenchmarkingagentskills} confirm these risks empirically: curated libraries contain skills that hurt performance, and larger libraries yield diminishing returns.

A complementary line of work trains the policy via agentic reinforcement learning (Agentic RL) against environment reward. Pairing skills or other accumulated memory with RL is a natural combination: skills focus exploration with behavioral priors, while RL shapes the policy through interaction. Recent skill-augmented RL injects retrieved skills into the agent's context each rollout~\citep{skillrl}; related threads close an offline-to-online distillation loop~\citep{evolver} or run policy updates directly over episodic memory~\citep{memrl}. Yet these methods largely treat the library as append-only, leaving the agent's context polluted by low-quality skills, bloated by unbounded accumulation, and misled by once-useful skills that turn counterproductive as the policy improves, a failure mode we call \textit{delayed obsolescence}: a skill whose fitness once reached the stable state but later drops below the retirement threshold as the policy matures.

\begin{figure*}[t] 
    \centering
    \includegraphics[width=\textwidth]{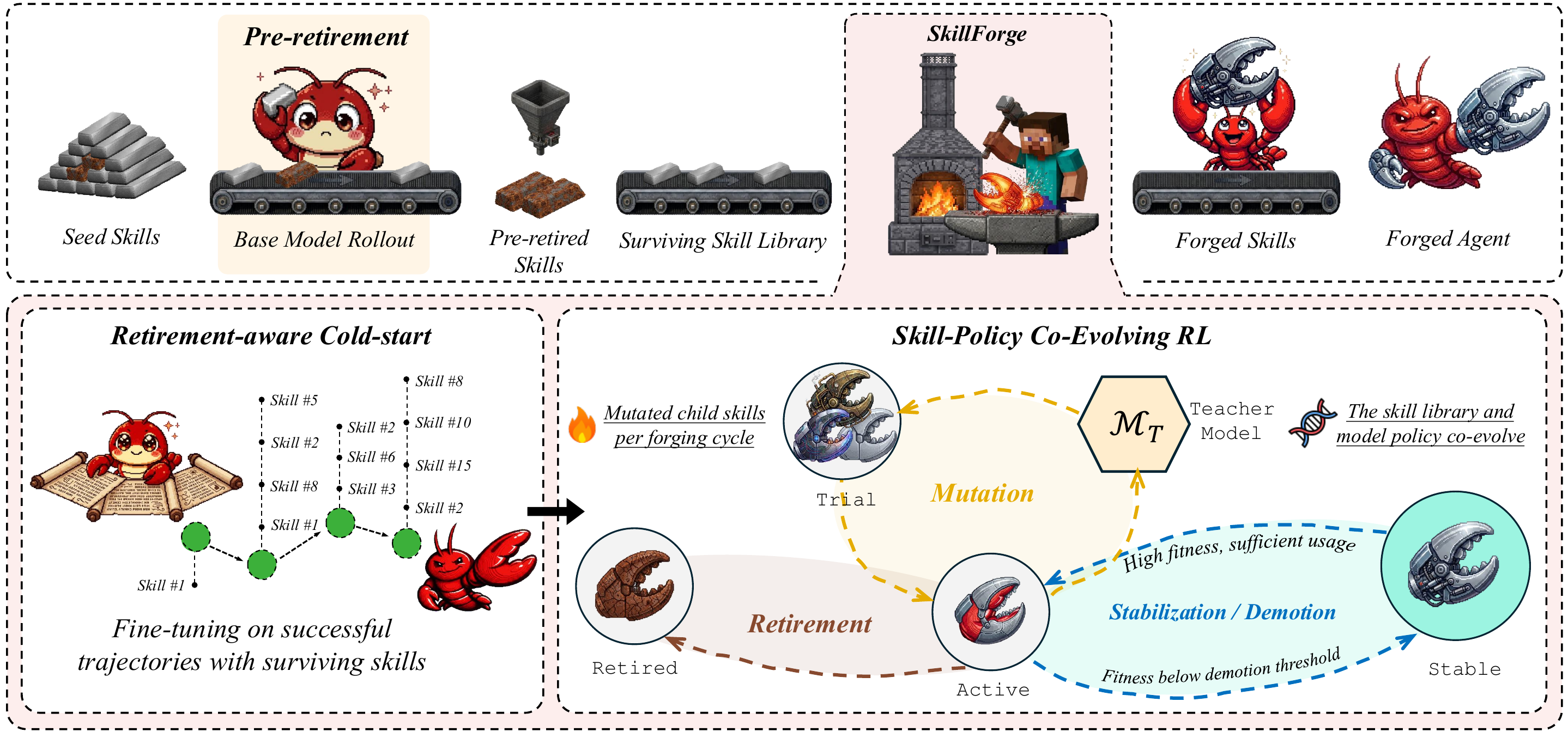}
    \caption{Overview of \method. \textbf{Top}: seed skills are pre-retired under the base model's own rollouts; surviving fitness-verified skills seed retirement-aware cold-start and skill-policy co-evolving RL, yielding a trained agent with an adapted skill library. \textbf{Bottom-left}: cold-start fine-tunes on successful trajectories with surviving skills. \textbf{Bottom-right}: skills and model co-evolve through the fitness-driven lifecycle across \texttt{trial}, \texttt{active}, \texttt{stable}, and \texttt{retired} via selective retirement, stabilization, demotion, and LLM-guided mutation.}
    \label{fig:overview}
\end{figure*}

We propose \method, which treats skill library management as an active \textit{forging} process rather than passive accumulation. Starting from a seed skill library distilled from trajectories, as illustrated in Figure~\ref{fig:overview}, \method runs in three stages: \textbf{(i)} pre-retire seed skills that fail under the base model's own rollouts, \textbf{(ii)} cold-start fine-tuning on the surviving fitness-verified skills, and \textbf{(iii)} switch to RL training where skills and model co-evolve each iteration. Each cycle reads runtime fitness and moves every skill along the skill lifecycle \texttt{trial}, \texttt{active}, \texttt{stable}, \texttt{retired}, with borderline-fitness skills rewritten by an LLM-guided mutation operator. Skills and model co-evolve, keeping the active set compact and aligned with the agent's capability. 
We contribute the following:

\begin{itemize}
    \item \textbf{\method and the skill lifecycle}: We propose \method, a fitness-driven skill lifecycle augmented with LLM-guided mutation that moves each skill through \texttt{trial}, \texttt{active}, \texttt{stable}, and \texttt{retired} states so that skills and model co-evolve, replacing the append-only accumulation of prior skill-augmented RL.
    \item \textbf{\skillfurnace dataset}: We release \skillfurnace, a three-part bundle of 5k+ annotated records spanning retirement-filtered SFT trajectories, evolved skill libraries with fitness annotations, and retirement events with human-annotated failure categories.
    \item \textbf{Empirical validation}: Across three interactive agent benchmarks spanning embodied control, web navigation, and Search-Augmented QA, \method surpasses the strongest baseline on every environment while keeping the library compact, delivering up to 7.8\% relative improvement in aggregate success rate.
\end{itemize}

\section{Related Work}
\label{sec:related}

\paragraph{Experience accumulation for LLM agents.}
Early work took the form of in-context reflection: self-critiquing the current output~\citep{reflexion,selfrefine}, and subsequently mining cross-trajectory insights~\citep{clin,expel}. In parallel, skill-library methods emerged: \citet{voyager} introduced an executable-code library for open-ended exploration, followed by reusable task workflows~\citep{awm}, and more recently procedural memory mixing step-level instructions with script-level abstractions~\citep{memp}. A related thread autonomously discovers new skills for the agent to practice~\citep{pae,exif}. On the RL side, memory-augmented methods couple such repositories to the training loop: earlier work used external memory to support tool-assisted planning~\citep{agentq,digirlx}, and later closed an offline-to-online loop between distillation and retrieval~\citep{evolver}. Most recently, \citet{skillrl} combined skill libraries with RL training, injecting retrieved skills into the context during each rollout. In most of them the repository is effectively monotone: entries persist once added and cannot be corrected as the policy evolves. \method instead treats the library as a dynamic population under fitness-driven selection, retiring low-fitness skills and reshaping borderline ones via LLM-guided mutation in step with the online RL optimizer.

\paragraph{LLM-guided evolutionary optimization.}
\citet{pbt} introduced population-based training, applying fitness-based selection and mutation over the hyperparameter schedules of concurrent workers. The same evaluate-select-mutate template has since been instantiated with LLMs: LLM-as-optimizer rewrites evolve a population of natural-language instructions against a fixed held-out metric~\citep{opro,evoprompt}, self-referential variants let instructions evolve each other~\citep{promptbreeder}, and quality-diversity searches pair a pretrained LLM with an evaluator to discover diverse high-performing programs~\citep{openelm,funsearch}. These methods all optimize static text artifacts against a stationary held-out metric in an offline loop. \method brings fitness-driven selection and mutation into LLM agentic reinforcement learning, where the evolving entities are \emph{skills} that augment the agent's context during interactive rollouts. Joint online training makes skill fitness inherently non-stationary, requiring a lifecycle that retires and reshapes skills as the policy evolves.

\section{Preliminaries}
\label{sec:prelim}

\subsection{Group Relative Policy Optimization}
\label{sec:grpo}

GRPO~\citep{grpo} estimates advantages from grouped samples without a separate value function. Given task description $d$ drawn from a training task distribution $\mathcal{D}_{\text{task}}$, a uniform distribution over the training tasks of each environment, the old policy $\pi_{\theta_\text{old}}$ samples $G$ episodes $\{\tau_n\}_{n=1}^{G}$ with binary rewards $r_n = R(\tau_n) \in \{0,1\}$, and computes group-normalized advantages $\hat{A}_n = (r_n - \mu_G)/\sigma_G$. Let $\boldsymbol{y}_n = (y_{n,1}, \dots, y_{n,|\boldsymbol{y}_n|})$ denote the concatenation of action tokens produced autoregressively by the policy during episode $\tau_n$; agent contexts and observations serve as conditioning inputs rather than generation targets, and $\boldsymbol{y}_n$ aligns with the step-level actions $\{a_t\}$ of Section~\ref{sec:skill_context}. Token-level importance ratios are $\rho_{n,l} = \pi_\theta(y_{n,l}\mid c_{n,l}) / \pi_{\theta_\text{old}}(y_{n,l}\mid c_{n,l})$, where $c_{n,l}$ denotes the left context of token $y_{n,l}$, comprising all prior agent contexts, observations, and earlier action tokens of the episode. The index $n$ indexes episodes within a GRPO group and is distinct from the skill index $i$ used in Section~\ref{sec:method}. The policy is updated by maximizing:
\begin{equation}
\begin{aligned}
    \mathcal{J}_{\text{GRPO}}(\theta) = {}& \mathbb{E}_{d \sim \mathcal{D}_{\text{task}},\,\{\tau_n\}}\!\left[\frac{1}{G}\sum_{n=1}^{G}\frac{1}{|\boldsymbol{y}_n|}\sum_{l=1}^{|\boldsymbol{y}_n|} \min\!\big(\rho_{n,l}\,\hat{A}_n,\;\operatorname{clip}(\rho_{n,l},\,1\!-\!\epsilon,\,1\!+\!\epsilon)\,\hat{A}_n\big)\right] \\
    & - \beta\,D_{\text{KL}}(\pi_\theta \| \pi_\text{ref}),
\end{aligned}
    \label{eq:grpo}
\end{equation}
with token-level clipping and a trajectory-level KL penalty against the reference policy.

\subsection{Skill-Augmented RL}
\label{sec:skill_context}

SkillRL~\citep{skillrl} augments GRPO with a textual skill library $\mathcal{S}$, where each skill $s = (v, b, \omega)$ comprises a name $v$, a behavioral principle $b$, and an applicability condition $\omega$. Given task description $d$ from Eq.~\eqref{eq:grpo}, a retrieval function $\phi(d, \mathcal{S})$ selects the top-$K$ skills whose condition $\omega$ matches the task-type label of $d$ and concatenates them into the agent's context:
\begin{equation}
    p_t = d \oplus \phi(d, \mathcal{S}) \oplus h_t \oplus o_t,
    \label{eq:context}
\end{equation}
where $h_t$ is a sliding window of the last $W$ observation--action pairs, $o_t$ is the current observation, and $\oplus$ denotes string concatenation. At each step $t$, an action $a_t \sim \pi_\theta(\cdot \mid p_t)$ is sampled, and the trajectory $\tau = (p_1, a_1, o_2, \ldots, p_T, a_T)$ decomposes into context--action pairs $\{(p_t, a_t)\}_{t=1}^{T}$ used as SFT supervision units. Each episode receives a binary reward $R(\tau) \in \{0, 1\}$. Initially, skills are distilled from trajectories via OpenAI o3~\citep{o3}, and a cold-start fine-tuning checkpoint $\pi_{\theta_\text{sft}}$ serves as both RL initialization and reference policy $\pi_\text{ref}$.

\section{\method}
\label{sec:method}

As illustrated in Figure~\ref{fig:overview}, \method runs in three stages: \textbf{(i)} pre-retirement evaluation filters low-fitness seed skills, \textbf{(ii)} retirement-aware cold-start fine-tuning on the surviving library, and \textbf{(iii)} agentic RL with a fitness-driven skill lifecycle where skills and policy co-evolve each iteration. This design addresses the non-stationarity of skill fitness under online RL: as the policy matures, once-useful skills may slip below the retirement threshold, \textit{delayed obsolescence} formalized in Eq.~\eqref{eq:nonstationary}.

\subsection{\skillfurnace Initialization}
\label{sec:presft}

We evaluate the initial skill library $\mathcal{S}_0$ released by SkillRL~\citep{skillrl} with base model $\pi_{\theta_0}$, which performs $M$ rollout episodes with skills injected. For each $s_i \in \mathcal{S}_0$ we accumulate usage $U_i^{(0)}$ and success $C_i^{(0)}$. Skills whose proto-fitness, their success rate under the base model's own rollouts, falls below a conservative threshold $\delta_{\text{pre}}$ and that have been used at least $N_{\text{min}}$ times are pre-retired:
\begin{equation}
    \hat{f}_i = \frac{C_i^{(0)}}{U_i^{(0)}}, \qquad \mathcal{S}_{\text{retire}}^{(0)} = \{s_i \in \mathcal{S}_0 \mid \hat{f}_i < \delta_{\text{pre}} \wedge U_i^{(0)} \geq N_{\text{min}}\}.
    \label{eq:proto_fitness}
\end{equation}
The surviving library $\mathcal{S}^{(0)} = \mathcal{S}_0 \setminus \mathcal{S}_{\text{retire}}^{(0)}$ seeds both the cold-start fine-tuning data and the RL phase described in Section~\ref{sec:scheduling}. Let $\mathcal{T}_{\text{eval}}$ denote the trajectories from this evaluation, and $\phi_{\text{eval}}(d, \mathcal{S}_0)$ the retrieval $\phi$ from Eq.~\eqref{eq:context} on the full library $\mathcal{S}_0$ before pre-retirement. We construct the SFT dataset from successful trajectories:
\begin{equation}
    \mathcal{D}_{\text{SFT}} = \{(p_t, a_t) \mid \tau \in \mathcal{T}_{\text{eval}},\, (p_t, a_t) \in \tau,\, R(\tau)=1,\, \phi_{\text{eval}}(d_\tau, \mathcal{S}_0) \cap \mathcal{S}_{\text{retire}}^{(0)} = \emptyset \},
    \label{eq:sft_data}
\end{equation}
where $d_\tau$ is the task description of $\tau$ and $R(\tau)$ its binary reward; the filter excludes trajectories that relied on subsequently retired skills. Fine-tuning the base model on $\mathcal{D}_{\text{SFT}}$ via standard cross-entropy
\begin{equation}
    \mathcal{L}_{\text{SFT}}(\theta) = -\mathbb{E}_{(p_t, a_t) \sim \mathcal{D}_{\text{SFT}}} \left[\log \pi_\theta(a_t \mid p_t)\right],
    \label{eq:sft_loss}
\end{equation}
produces $\pi_{\theta_\text{sft}}$, which serves as both the RL initialization and the reference policy $\pi_\text{ref}$.

The filtered $\mathcal{D}_{\text{SFT}}$, the surviving library $\mathcal{S}^{(0)}$, and human-annotated retirement events from previously recorded training runs together form \skillfurnace, our released dataset. As summarized in Table~\ref{tab:skillfurnace}, retired skills peak at an average fitness of 0.67, close to the stabilization threshold $\delta_{\text{stable}}$ of 0.7, before dropping by 0.28 at retirement, and 57.2\% first reach the \texttt{stable} state before this decay: a direct empirical signature of \textit{delayed obsolescence}. Appendix~\ref{app:skillfurnace} gives per-component details.

\begin{table}[t] 
\caption{\skillfurnace dataset statistics. The top block reports per-component record counts across SFT trajectories, library snapshots, and retirement events; the bottom block reports detailed retirement-event metrics. \textbf{Total records (\#)} is summed across components within each environment.}
\label{tab:skillfurnace}
\centering
\renewcommand{\arraystretch}{1.05}
\setlength{\aboverulesep}{2pt}
\setlength{\belowrulesep}{2pt}
\resizebox{\columnwidth}{!}{%
\setlength{\tabcolsep}{18pt}
\begin{tabular}{lcccc}
\toprule
\textbf{Statistic} & \textbf{ALFWorld} & \textbf{WebShop} & \textbf{Search-Aug. QA} & \textbf{Total} \\
\midrule
\multicolumn{5}{l}{\emph{Dataset scale}} \\
SFT trajectories (\#)         & 412 & 348 & 253 & 1{,}013 \\
Library snapshot records (\#) & 1{,}512 & 1{,}683 & 1{,}326 & 4{,}521 \\
Retirement events (\#)        & 138 & 121 & 59 & 318 \\
\textbf{Total records (\#)}   & \textbf{2{,}062} & \textbf{2{,}152} & \textbf{1{,}638} & \textbf{5{,}852} \\
\midrule
\multicolumn{5}{l}{\emph{Retirement event statistics}} \\
Avg.\ peak fitness            & 0.696 & 0.709 & 0.515 & 0.667 \\
Avg.\ fitness drop from peak  & 0.294 & 0.279 & 0.239 & 0.278 \\
Avg.\ lifespan (steps)        & 38.8 & 31.7 & 49.4 & 38.0 \\
Stable before retirement (\%) & 60.1 & 58.7 & 47.5 & 57.2 \\
\bottomrule
\end{tabular}
}
\end{table}

\subsection{Skill Fitness Tracking}
\label{sec:fitness}

For each skill $s_i$, we maintain a usage count $U_i$ (episodes in which $s_i$ was retrieved) and a success count $C_i$ (episodes that succeeded). At the start of RL, the counters are initialized from the preliminary evaluation phase: $C_i \leftarrow C_i^{(0)}$ and $U_i \leftarrow U_i^{(0)}$ for each surviving skill $s_i \in \mathcal{S}^{(0)}$, so proto-fitness information from Section~\ref{sec:presft} is preserved rather than discarded. The fitness is:
\begin{equation}
    f_i = \begin{cases}
        C_i / U_i & \text{if } U_i \geq N_{\text{warm}} \\
        f_{\text{default}} & \text{otherwise}
    \end{cases}
    \label{eq:fitness}
\end{equation}
where $N_{\text{warm}}$ is a minimum usage count before trusting the empirical success rate and $f_{\text{default}}$ is a neutral prior for new skills. We use $f_i^{(k)}$ for the running estimate at training step $k$ when dependence on $k$ matters, as formalized in Section~\ref{sec:scheduling}; for any skill with $U_i^{(0)} \geq N_{\text{warm}}$ we have $f_i^{(0)} = \hat{f}_i$, so the proto-fitness from Eq.~\eqref{eq:proto_fitness} is the runtime fitness at step 0, reused rather than recomputed.

Credit is assigned at the episode level: for each episode with trajectory $\tau$ and retrieved skill set $\mathcal{S}_\tau = \phi(d, \mathcal{S})$, all $K$ retrieved skills receive the binary reward signal:
\begin{equation}
    U_i \leftarrow U_i + \mathbb{1}[s_i \in \mathcal{S}_\tau], \quad C_i \leftarrow C_i + \mathbb{1}[s_i \in \mathcal{S}_\tau] \cdot R(\tau),
    \label{eq:credit}
\end{equation}
where $\mathbb{1}[\cdot]$ denotes the indicator function.
\subsection{Skill Lifecycle Management}
\label{sec:lifecycle}

Based on accumulated fitness, we define a \emph{lifecycle update rule} $\ell$ that assigns each skill one of four states $\{\texttt{trial}, \texttt{active}, \texttt{stable}, \texttt{retired}\}$; the active library $\mathcal{S}$ retains only non-retired skills. Transitions depend on fitness thresholds $\delta_{\text{retire}} < \delta_{\text{demote}} < \delta_{\text{stable}}$, where $\delta_{\text{demote}}$ lies strictly between the retirement and stabilization thresholds to provide hysteresis so that a stable skill whose fitness slips must first be demoted to \texttt{active} before becoming eligible for retirement. Transitions also depend on a mutation range $[\delta_{\text{low}}, \delta_{\text{high}}]$, minimum usage counts $N_{\text{promote}}$ and $N_{\text{stable}}$, and a generation-aware retirement threshold:
\begin{equation}
    N_{\text{retire}}^{(g_i)} = \begin{cases}
        N_{\text{protect}} & \text{if } g_i = 0 \\
        N_{\text{retire}}  & \text{otherwise}
    \end{cases}
    \label{eq:retire_threshold}
\end{equation}
where $g_i$ is the mutation generation; distilled ($g_i = 0$) skills require a higher usage bar $N_{\text{protect}} > N_{\text{retire}}$ before they can be retired. The lifecycle update rule is then:
\begin{equation}
    \ell(s_i) \leftarrow \begin{cases}
        \texttt{active}  & \text{if } \ell(s_i) = \texttt{trial} \wedge U_i \geq N_{\text{promote}} \\
        \texttt{stable}  & \text{if } \ell(s_i) = \texttt{active} \wedge f_i \geq \delta_{\text{stable}} \wedge U_i \geq N_{\text{stable}} \\
        \texttt{active}  & \text{if } \ell(s_i) = \texttt{stable} \wedge f_i < \delta_{\text{demote}} \\
        \texttt{retired} & \text{if } \ell(s_i) = \texttt{active} \wedge f_i < \delta_{\text{retire}} \wedge U_i \geq N_{\text{retire}}^{(g_i)}
    \end{cases}
    \label{eq:transitions}
\end{equation}
If none applies, for example an active skill with $\delta_{\text{retire}} \leq f_i < \delta_{\text{stable}}$, $\ell(s_i)$ is left unchanged; Appendix~\ref{app:hyperparams} states the well-definedness and per-cycle ordering used by Algorithm~\ref{alg:forging}. Only \texttt{active} skills are eligible for retirement; \texttt{trial} skills must first accumulate $N_{\text{promote}}$ usages and promote to \texttt{active}, and \texttt{stable} skills must first be demoted. This two-gate design prevents premature retirement of newly created skills and shields proven skills from transient fitness dips. The mutation generation $g_i$ is 0 for skills distilled into $\mathcal{S}_0$ and is incremented by 1 at each mutation.

Initial skills surviving pre-retirement enter as \texttt{stable} if $\hat{f}_i \geq \delta_{\text{stable}}$, or as \texttt{active} otherwise; newly generated skills enter as \texttt{trial}. The lifecycle operates through promotion, demotion, retirement, stabilization, and mutation, per Algorithm~\ref{alg:forging}. Because demotion precedes retirement, a skill can be demoted and retired in one cycle, so once-stable skills can be retired as the policy evolves.

\textbf{Mutation} reshapes \texttt{active} skills whose fitness lies in a moderate range $[\delta_{\text{low}}, \delta_{\text{high}}]$ and which have been used at least $N_{\text{mutate}}$ times, so that the fitness estimate used to select mutation parents is informative. From the resulting candidate pool $\mathcal{M}_{\text{cand}}$, we select up to $K_{\text{mutate}}$ parents via inverse-fitness-weighted sampling ($P(s_i) \propto 1 - f_i$), collect their recent failure trajectories $\mathcal{T}_{\text{fail}}$, and query our teacher model ($\mathcal{M}_T$) to generate improved variants:
\begin{equation}
\begin{aligned}
    \mathcal{M}_{\text{cand}} &= \{s_i \in \mathcal{S} \mid \ell(s_i) = \texttt{active} \wedge \delta_{\text{low}} \leq f_i \leq \delta_{\text{high}} \wedge U_i \geq N_{\text{mutate}}\}, \\
    s_{\text{child}} &= \mathcal{M}_T\big(\mathcal{P}_\mu(s_{\text{parent}},\, f_{\text{parent}},\, \mathcal{T}_{\text{fail}})\big),
\end{aligned}
    \label{eq:mutation}
\end{equation}
where $\mathcal{P}_\mu$ assembles the parent skill's content, fitness statistics, and failure examples into a structured template. Each child enters with $\ell(s_{\text{child}}) = \texttt{trial}$ and generation $g_{\text{child}} = g_{\text{parent}} + 1$. The inverse-fitness weighting $P(s_i) \propto 1-f_i$ biases selection toward skills with more room to improve without concentrating mutation on a single outlier. We instantiate $\mathcal{M}_T$ from a different model family than the trained policy, since \citet{li2026skillsbenchbenchmarkingagentskills} report that self-generated skills provide negligible or negative benefit compared with externally authored ones.

\subsection{Training Integration}
\label{sec:scheduling}

As given in Algorithm~\ref{alg:forging}, a forging cycle runs every $E$ validation epochs, with per-cycle budgets $K_{\text{retire}}$ and $K_{\text{mutate}}$ and a total skill cap $S_{\text{max}}$ bounding computational cost.

\begin{algorithm}[t]
\caption{\method Forging Cycle}
\label{alg:forging}
\begin{algorithmic}[1]
\REQUIRE Skill library $\mathcal{S}$, fitness scores $\{f_i\}$, usage counts $\{U_i\}$, failed trajectories $\mathcal{T}_{\text{fail}}$
\ENSURE Updated skill library $\mathcal{S}$
\FOR{$s_i \in \mathcal{S}$ where $\ell(s_i) = \texttt{trial} \wedge U_i \geq N_{\text{promote}}$}
    \STATE $\ell(s_i) \leftarrow \texttt{active}$ \hfill $\triangleright$ Promote: trial $\rightarrow$ active
\ENDFOR
\FOR{$s_i \in \mathcal{S}$ where $\ell(s_i) = \texttt{stable} \wedge f_i < \delta_{\text{demote}}$}
    \STATE $\ell(s_i) \leftarrow \texttt{active}$ \hfill $\triangleright$ Demote: stable $\rightarrow$ active
\ENDFOR
\STATE $\mathcal{R} \leftarrow \{s_i \in \mathcal{S} \mid \ell(s_i) = \texttt{active} \wedge f_i < \delta_{\text{retire}} \wedge U_i \geq N_{\text{retire}}^{(g_i)}\}$ \hfill $\triangleright$ Retirement candidates
\STATE $\mathcal{S} \leftarrow \mathcal{S} \setminus \text{bottom-}m(\mathcal{R}),\ m = \min(K_{\text{retire}}, |\mathcal{R}|)$ \hfill $\triangleright$ $\text{bottom-}m(\mathcal{R})$: $m$ skills in $\mathcal{R}$ with lowest $f_i$
\FOR{$s_i \in \mathcal{S}$ where $\ell(s_i) = \texttt{active} \wedge f_i \geq \delta_{\text{stable}} \wedge U_i \geq N_{\text{stable}}$}
    \STATE $\ell(s_i) \leftarrow \texttt{stable}$ \hfill $\triangleright$ Stabilize: active $\rightarrow$ stable
\ENDFOR
\STATE $\mathcal{M}_{\text{cand}} \leftarrow \{s_i \in \mathcal{S} \mid \ell(s_i) = \texttt{active} \wedge \delta_{\text{low}} \leq f_i \leq \delta_{\text{high}} \wedge U_i \geq N_{\text{mutate}}\}$ \hfill $\triangleright$ Mutation pool
\STATE Select $K_{\text{mutate}}$ parents from $\mathcal{M}_{\text{cand}}$; $P(s_i) \propto 1 - f_i$ \hfill $\triangleright$ Lower fitness $\rightarrow$ higher chance
\FOR{each selected $s_{\text{parent}}$}
    \STATE $s_{\text{child}} \leftarrow \mathcal{M}_T(\mathcal{P}_\mu(s_{\text{parent}}, f_{\text{parent}}, \mathcal{T}_{\text{fail}}))$ \hfill $\triangleright$ LLM-guided mutation
    \STATE $\mathcal{S} \leftarrow \mathcal{S} \cup \{s_{\text{child}}\}$; $\ell(s_{\text{child}}) \leftarrow \texttt{trial}$; $g_{\text{child}} \leftarrow g_{\text{parent}} + 1$ \hfill $\triangleright$ Child enters as trial
\ENDFOR
\WHILE{$|\mathcal{S}| > S_{\text{max}}$ \AND $\exists s_i \in \mathcal{S}$ with $\ell(s_i)=\texttt{active}$}
    \STATE $\mathcal{S} \leftarrow \mathcal{S} \setminus \{\arg\min_{s_i:\, \ell(s_i)=\texttt{active}} f_i\}$ \hfill $\triangleright$ Enforce size bound
\ENDWHILE
\RETURN $\mathcal{S}$
\end{algorithmic}
\end{algorithm}

Making the non-stationarity introduced at the start of Section~\ref{sec:method} explicit, the fitness estimate at training step $k$ depends on the policy trajectory:
\begin{equation}
    f_i^{(k)} = \frac{\sum_{j \leq k} \mathbb{1}[s_i \in \mathcal{S}_{\tau^{(j)}}] \cdot R(\tau^{(j)})}{\sum_{j \leq k} \mathbb{1}[s_i \in \mathcal{S}_{\tau^{(j)}}]}, \quad \tau^{(j)} \sim \pi_{\theta_j}(\cdot \mid p^{(j)}),
    \label{eq:nonstationary}
\end{equation}
where $p^{(j)}$ denotes the agent's context at step $j$, constructed via Eq.~\eqref{eq:context} with the library $\mathcal{S}^{(j)}$. The $j=0$ term covers the pre-retirement rollouts under $\pi_{\theta_0}$, consistent with the counter initialization in Section~\ref{sec:fitness} and yielding $f_i^{(0)} = \hat{f}_i$. As \citet{li2026rethinking} report, on-policy formulations avoid the exposure bias of off-policy alternatives, justifying the non-stationarity above.
At each step $k$, the policy is updated via Eq.~\eqref{eq:grpo} conditioned on the current library, and the forging cycle updates the library based on accumulated fitness:
\begin{equation}
    \theta_{k+1} = \arg\max_\theta\, \mathcal{J}_\text{GRPO}(\theta;\, \mathcal{S}^{(k)}), \qquad \mathcal{S}^{(k+1)} = \operatorname{Forge}\big(\mathcal{S}^{(k)},\, \{f_i^{(k)}\}\big),
    \label{eq:coupled}
\end{equation}
where $\operatorname{Forge}$ denotes a full forging cycle. Skills and policy thus co-evolve: as $\theta$ improves, fitness estimates shift, and the forging cycle retires skills whose utility has diminished. \textit{Delayed obsolescence} is then the formal event $\exists k_1 < k_2:\ f_i^{(k_1)} \geq \delta_{\text{stable}} \wedge f_i^{(k_2)} < \delta_{\text{retire}}$.

\section{Experiments}
\label{sec:experiments}

\subsection{Experimental Setup}
\label{sec:setup}

\paragraph{Environments.}
Three interactive agent benchmarks are used: ALFWorld~\citep{alfworld}, WebShop~\citep{webshop}, and Search-Augmented QA~\citep{searchr1}, with the Search-Augmented QA setting spanning NQ~\citep{nq}, TriviaQA~\citep{triviaqa}, PopQA~\citep{popqa}, HotpotQA~\citep{hotpotqa}, 2WikiMultiHopQA~\citep{twowikimhqa}, MuSiQue~\citep{musique}, and Bamboogle~\citep{bamboogle}.

\paragraph{Baselines.}
Baselines fall into three categories. \textbf{Context and memory methods} rely on in-context reasoning or external experience without updating the policy: ReAct~\citep{react}, Reflexion~\citep{reflexion}, Mem0~\citep{mem0}, ExpeL~\citep{expel}, MemP~\citep{memp}, and SimpleMem~\citep{simplemem}. \textbf{RL-based methods} optimize the policy via group-relative advantage estimation: RLOO~\citep{rloo} and GRPO~\citep{grpo}. \textbf{Memory-augmented RL methods} couple persistent memory with policy optimization and form the most direct comparison: MemRL~\citep{memrl}, EvolveR~\citep{evolver}, SkillRL~\citep{skillrl}, and two additional combinations, Mem0+GRPO and SimpleMem+GRPO, implemented by adding GRPO optimization to the respective memory mechanisms. For Search-Augmented QA the comparison additionally includes Search-o1~\citep{searcho1}, R1-Instruct~\citep{deepseekr1}, Search-R1~\citep{searchr1}, ZeroSearch~\citep{zerosearch}, and StepSearch~\citep{stepsearch}. GPT-4o~\citep{gpt4o} and Gemini-2.5-Pro~\citep{gemini25} are reported as frontier reference points.

\begin{table*}[t]
\caption{Performance on ALFWorld and WebShop (\%). WebShop \textbf{Score} is the partial-credit reward; \textbf{Succ.}\ is binary task success. \best{Bold}: best; \second{underline}: second best.}
\label{tab:main1}
\centering
\renewcommand{\arraystretch}{1.1}
\setlength{\aboverulesep}{2pt}
\setlength{\belowrulesep}{2pt}
\resizebox{\textwidth}{!}{%
\setlength{\tabcolsep}{4pt}
\begin{tabular}{l ccccccc cc}
\toprule
\multirow{2}{*}{\raisebox{-3pt}{\textbf{Method}}}
  & \multicolumn{7}{c}{\textbf{ALFWorld}}
  & \multicolumn{2}{c}{\textbf{WebShop}} \\
\cmidrule(lr){2-8} \cmidrule(lr){9-10}
  & \textbf{Pick} & \textbf{Look} & \textbf{Clean} & \textbf{Heat} & \textbf{Cool} & \textbf{Pick2} & \textbf{All}
  & \textbf{Score} & \textbf{Succ.} \\
\midrule
\multicolumn{10}{c}{\textbf{Closed-source LLMs}} \\
\midrule
GPT-4o & 75.3 & 60.8 & 31.2 & 56.7 & 21.6 & 49.8 & 48.0 & 31.8 & 23.7 \\
Gemini-2.5-Pro & 92.8 & 63.3 & 62.1 & 69.0 & 26.6 & 58.7 & 60.3 & 42.5 & 35.9 \\
\midrule
\multicolumn{10}{c}{\textbf{Base Model}} \\
\midrule
Vanilla & 33.4 & 21.6 & 19.3 & 6.9 & 2.8 & 3.2 & 14.8 & 26.4 & 7.8 \\
\midrule
\multicolumn{10}{c}{\textbf{Context / Memory Methods}} \\
\midrule
ReAct & 48.5\up{15.1} & 35.4\up{13.8} & 34.3\up{15.0} & 13.2\up{6.3} & 18.2\up{15.4} & 17.6\up{14.4} & 31.2\up{16.4} & 46.2\up{19.8} & 19.5\up{11.7} \\
Reflexion & 62.0\up{28.6} & 41.6\up{20.0} & 44.9\up{25.6} & 30.9\up{24.0} & 36.3\up{33.5} & 23.8\up{20.6} & 42.7\up{27.9} & 58.1\up{31.7} & 28.8\up{21.0} \\
Mem0 & 54.0\up{20.6} & 55.0\up{33.4} & 26.9\up{7.6} & 36.4\up{29.5} & 20.8\up{18.0} & 7.7\up{4.5} & 33.6\up{18.8} & 23.9\dn{2.5} & 2.0\dn{5.8} \\
ExpeL & 21.0\dn{12.4} & 67.0\up{45.4} & 55.0\up{35.7} & 52.0\up{45.1} & 71.0\up{68.2} & 6.0\up{2.8} & 46.3\up{31.5} & 30.9\up{4.5} & 11.2\up{3.4} \\
MemP & 54.3\up{20.9} & 38.5\up{16.9} & 48.1\up{28.8} & 56.2\up{49.3} & 32.0\up{29.2} & 16.7\up{13.5} & 41.4\up{26.6} & 25.3\dn{1.1} & 6.4\dn{1.4} \\
SimpleMem & 64.5\up{31.1} & 33.3\up{11.7} & 20.0\up{0.7} & 12.5\up{5.6} & 33.3\up{30.5} & 3.8\up{0.6} & 29.7\up{14.9} & 33.2\up{6.8} & 8.6\up{0.8} \\
\midrule
\multicolumn{10}{c}{\textbf{RL-based Methods}} \\
\midrule
RLOO & 87.6\up{54.2} & \second{78.2}\up{56.6} & 87.3\up{68.0} & 81.3\up{74.4} & 71.9\up{69.1} & 48.9\up{45.7} & 75.5\up{60.7} & 80.3\up{53.9} & 65.7\up{57.9} \\
GRPO & 90.8\up{57.4} & 66.1\up{44.5} & 89.3\up{70.0} & 74.7\up{67.8} & 72.5\up{69.7} & 64.7\up{61.5} & 77.6\up{62.8} & 79.3\up{52.9} & 66.1\up{58.3} \\
\midrule
\multicolumn{10}{c}{\textbf{Memory-Augmented RL}} \\
\midrule
MemRL & 62.8\up{29.4} & 38.5\up{16.9} & 22.2\up{2.9} & 12.5\up{5.6} & 8.0\up{5.2} & 0.0\dn{3.2} & 21.4\up{6.6} & 29.5\up{3.1} & 9.2\up{1.4} \\
EvolveR & 64.9\up{31.5} & 33.3\up{11.7} & 46.4\up{27.1} & 13.3\up{6.4} & 33.3\up{30.5} & 33.3\up{30.1} & 43.8\up{29.0} & 42.5\up{16.1} & 17.6\up{9.8} \\
Mem0+GRPO & 78.1\up{44.7} & 54.8\up{33.2} & 56.1\up{36.8} & 31.0\up{24.1} & 65.0\up{62.2} & 26.9\up{23.7} & 54.7\up{39.9} & 58.1\up{31.7} & 37.5\up{29.7} \\
SimpleMem+GRPO & 89.5\up{56.1} & 36.3\up{14.7} & 60.0\up{40.7} & 50.0\up{43.1} & 64.9\up{62.1} & 26.3\up{23.1} & 62.5\up{47.7} & 67.8\up{41.4} & 46.9\up{39.1} \\
SkillRL & \second{97.9}\up{64.5} & 71.4\up{49.8} & \best{90.0}\up{70.7} & \second{90.0}\up{83.1} & \second{95.5}\up{92.7} & \second{87.5}\up{84.3} & \second{89.9}\up{75.1} & \second{85.2}\up{58.8} & \second{72.7}\up{64.9} \\
\midrule
\ours \best{\method} & \ours\best{98.2}\up{64.8} & \ours\best{86.6}\up{65.0} & \ours\second{89.8}\up{70.5} & \ours\best{93.7}\up{86.8} & \ours\best{95.6}\up{92.8} & \ours\best{88.4}\up{85.2} & \ours\best{92.4}\up{77.6} & \ours\best{87.8}\up{61.4} & \ours\best{78.4}\up{70.6} \\
\bottomrule
\end{tabular}
}
\end{table*}

\begin{figure*}[t]
    \centering
    \begin{subfigure}[t]{0.45\textwidth}
        \centering
        \includegraphics[width=\linewidth]{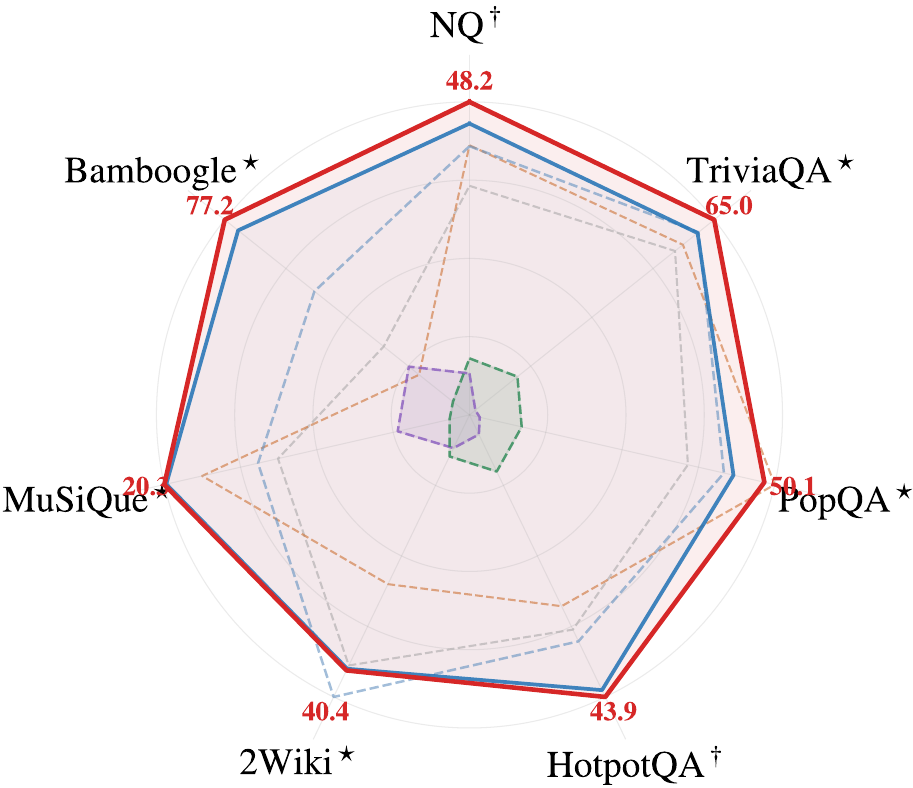}
        \caption{Search-Augmented QA comparison (\%).}
        \label{fig:searchqa_radar}
    \end{subfigure}
    \hfill
    \begin{subfigure}[t]{0.5\textwidth}
        \centering
        \includegraphics[width=\linewidth]{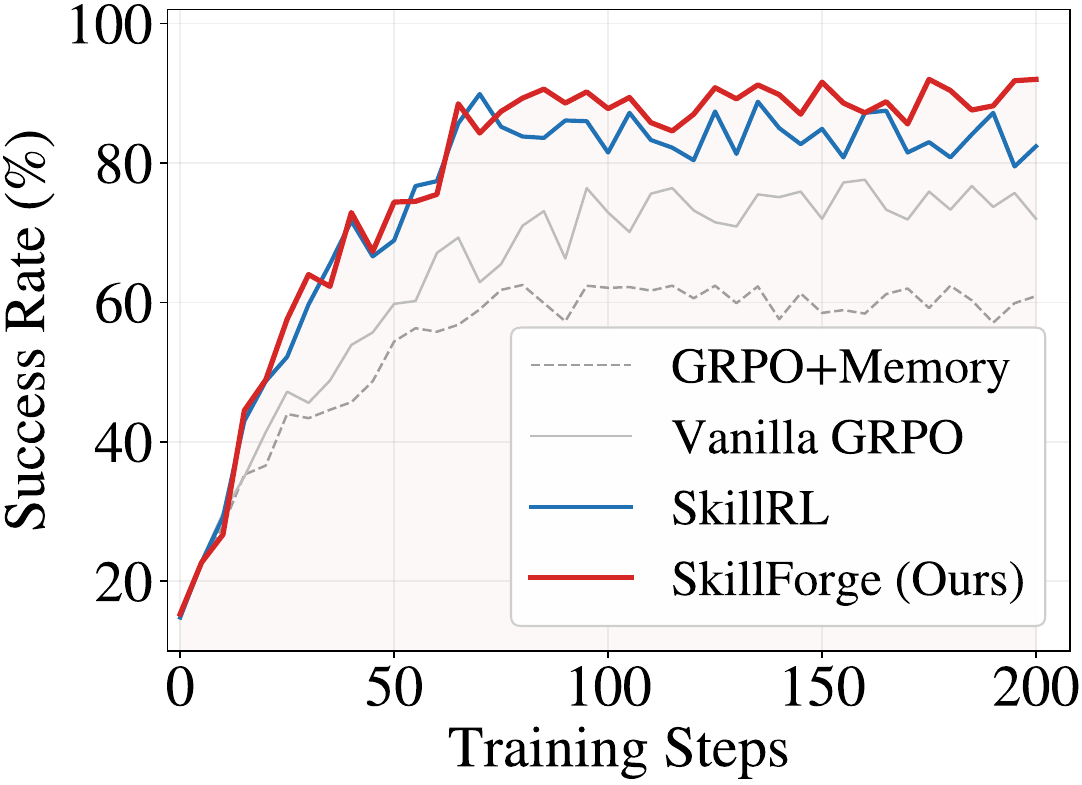}
        \caption{ALFWorld validation success rate across training.}
        \label{fig:curves}
    \end{subfigure}
    \caption{Main-results diagnostics.
    (a) Per-dataset accuracy on Search-Augmented QA; per-axis numbers in bold red report \method accuracy on each dataset.
    \textcolor{clrRetired}{\textbf{---}}: \method (Ours);
    \textcolor{clrStable}{\textbf{---}}: SkillRL;
    \textcolor{clrActive}{\textbf{- -}}: EvolveR;
    \textcolor{clrTrial}{\textbf{- -}}: ZeroSearch;
    \textcolor{clrGray}{\textbf{- -}}: Search-R1;
    \textcolor[HTML]{8B5FBF}{\textbf{- -}}: Search-o1;
    \textcolor[HTML]{2E8B57}{\textbf{- -}}: R1-Instruct.
    $^\dagger$In-domain; $^\star$out-of-domain.
    (b) \method achieves faster convergence and higher success rates than vanilla GRPO and memory-augmented RL.}
    \label{fig:main_results}
\end{figure*}

\paragraph{Implementation.}
All experiments are built on the \textit{verl} \citep{verbal} framework with GRPO as the policy optimizer. The base model is Qwen2.5-7B-Instruct~\citep{qwen}, first fine-tuned via the retirement-aware SFT described in Section~\ref{sec:presft} to produce environment-specific initializations. Training runs for 200 steps with a learning rate of $1 \times 10^{-6}$, clipping range $\epsilon = 0.2$, and a KL penalty coefficient $\beta = 0.001$. Rollout generation uses temperature $1.0$ with a group size of $G = 8$. The forging cycle executes every 10 training steps, with up to $K_{\text{mutate}} = 5$ mutations and $K_{\text{retire}} = 3$ retirements per cycle; the teacher model for mutation is Kimi-K2.5~\citep{kimik25}. Full hyperparameters, hardware details, and forging thresholds are listed in Appendix~\ref{app:hyperparams}.

\subsection{Main Results}
\label{sec:main_results}

As shown in Table~\ref{tab:main1}, \method achieves 92.4\% on ALFWorld and 78.4\% on WebShop, outperforming the closed-source GPT-4o and Gemini-2.5-Pro baselines by a wide margin and improving over SkillRL by 2.8\% relative on ALFWorld and 7.8\% relative on WebShop. The largest per-subtask relative improvement over SkillRL is on ALFWorld's Look at 21.3\%; the Clean subtask remains within 0.2\% of SkillRL, within sampling noise. On Search-Augmented QA, as illustrated in Figure~\ref{fig:main_results}\subref{fig:searchqa_radar}, \method reaches $48.7 \pm 0.3\%$ micro-averaged accuracy on the full test set, the highest overall among all methods; per-dataset results are in Appendix~\ref{app:searchqa_table}.

As illustrated in Figure~\ref{fig:main_results}\subref{fig:curves}, \method and SkillRL both climb from ${\sim}$60\% to ${\sim}$70\% between steps 50 and 65, but \method separates beyond step 75 as the forging lifecycle takes effect, reaching ${\sim}$90\% while SkillRL oscillates in the ${\sim}$80--88\% range over the rest of the training budget.
\begin{table}[t]
\caption{Ablation study of \method across three environments. ``Library'' reports the final non-retired skill count (coinciding with $|\mathcal{S}_0|$ for $-$forging/$-$mutation). Each row disables one forging stage: w/o forging removes the entire \method lifecycle, w/o pre-retirement keeps runtime stages but skips the pre-RL fitness filter, w/o retirement disables skill removal and size capping, and w/o mutation disables LLM-guided refinement. \best{Bold}: best.}
\label{tab:ablation}
\centering
\setlength{\aboverulesep}{2pt}
\setlength{\belowrulesep}{2pt}
\resizebox{\columnwidth}{!}{%
\setlength{\tabcolsep}{12pt}
\begin{tabular}{l cc cc cc}
\toprule
\multirow{2}{*}{\raisebox{-3pt}{\textbf{Configuration}}}
  & \multicolumn{2}{c}{\textbf{ALFWorld}}
  & \multicolumn{2}{c}{\textbf{WebShop}}
  & \multicolumn{2}{c}{\textbf{Search-Aug. QA}} \\
\cmidrule(lr){2-3} \cmidrule(lr){4-5} \cmidrule(lr){6-7}
  & \textbf{Library} & \textbf{Succ.\ (\%)}
  & \textbf{Library} & \textbf{Succ.\ (\%)}
  & \textbf{Library} & \textbf{Succ.\ (\%)} \\
\midrule
\ours \method & \ours 100 & \ours \best{92.4} & \ours 95 & \ours \best{78.4} & \ours 85 & \ours \best{48.7} \\
\quad w/o forging & 44 & 89.8\dn{2.6} & 66 & 72.6\dn{5.8} & 41 & 46.3\dn{2.4} \\
\quad w/o pre-retirement & 100 & 90.7\dn{1.7} & 95 & 74.9\dn{3.5} & 85 & 47.5\dn{1.2} \\
\quad w/o retirement & 132 & 91.4\dn{1.0} & 121 & 75.9\dn{2.5} & 111 & 47.9\dn{0.8} \\
\quad w/o mutation & 44 & 91.8\dn{0.6} & 66 & 76.8\dn{1.6} & 41 & 48.2\dn{0.5} \\
\bottomrule
\end{tabular}
}
\end{table}

\subsection{Ablation Study}
\label{sec:ablation}

As shown in Table~\ref{tab:ablation}, removing the entire \method lifecycle produces the largest relative drops on every environment, peaking at 7.4\% on WebShop. Pre-retirement is the single most impactful component, with per-stage relative drops of 1.8\%, 4.5\%, and 2.5\% on ALFWorld, WebShop, and Search-Augmented QA respectively. The single-component drops consistently sum to more than the joint w/o forging drop, consistent with sub-additive interactions among the three components.

Larger libraries do not imply better performance. Without retirement the library grows to over 130 skills on ALFWorld, yet \method outperforms it with a compact set of 100. On WebShop, enabling pre-retirement and mutation lifts the static $72.6\%$ baseline to $75.9\%$; adding retirement trims the library from 121 skills to 95 and closes the remaining $3.2\%$ relative gap to the full $78.4\%$, confirming that quality control, not scale, drives the gain.

\subsection{Forging Dynamics}
\label{sec:evo_dynamics}

\begin{figure}[t]
    \centering
    \begin{subfigure}[t]{0.325\textwidth}
        \includegraphics[width=\textwidth]{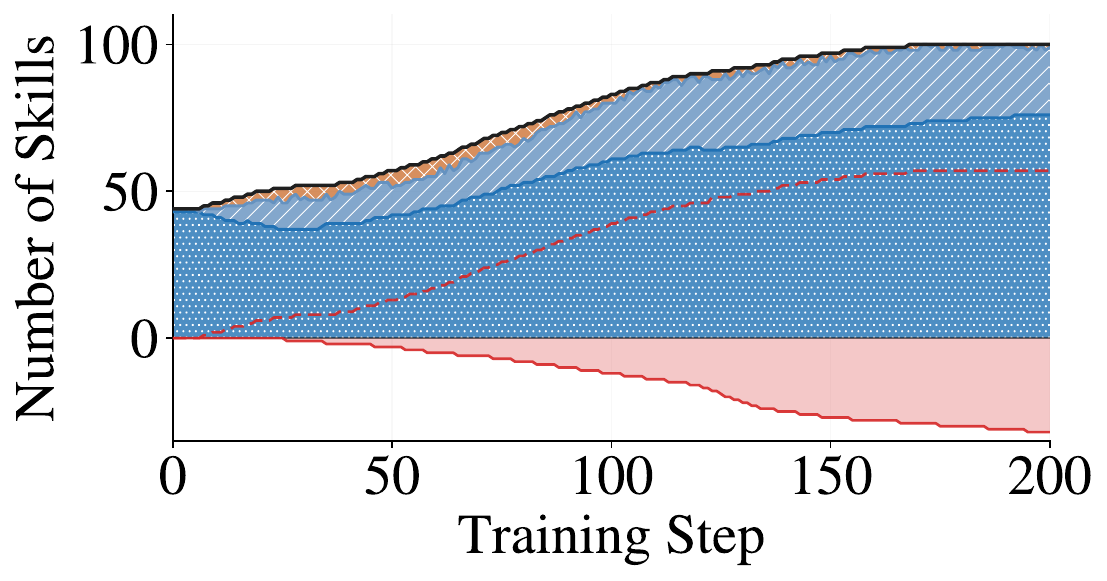}
        \caption{ALFWorld}
    \end{subfigure}
    \hfill
    \begin{subfigure}[t]{0.325\textwidth}
        \includegraphics[width=\textwidth]{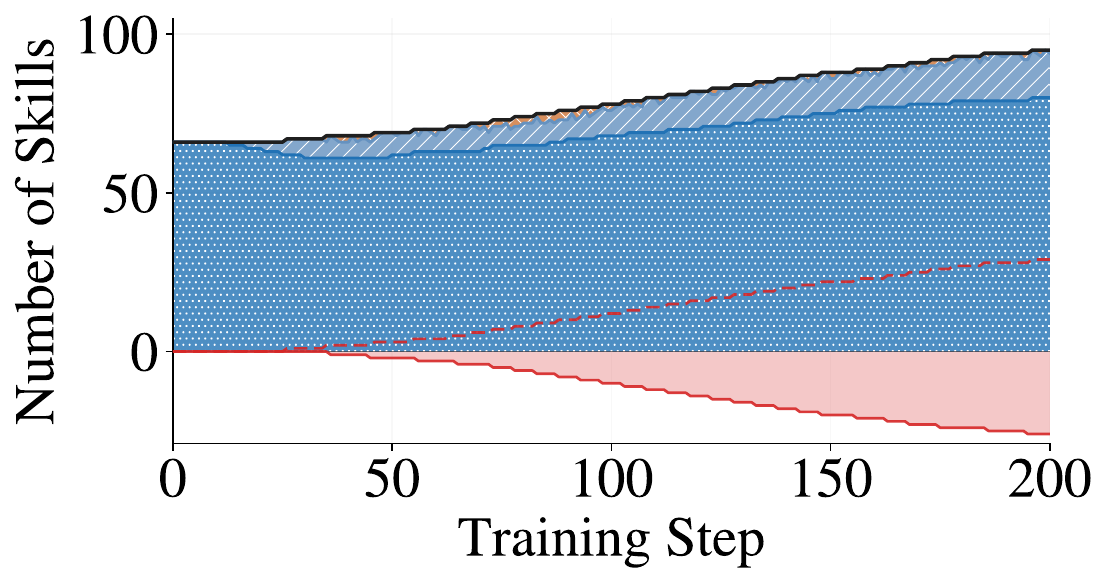}
        \caption{WebShop}
    \end{subfigure}
    \hfill
    \begin{subfigure}[t]{0.325\textwidth}
        \includegraphics[width=\textwidth]{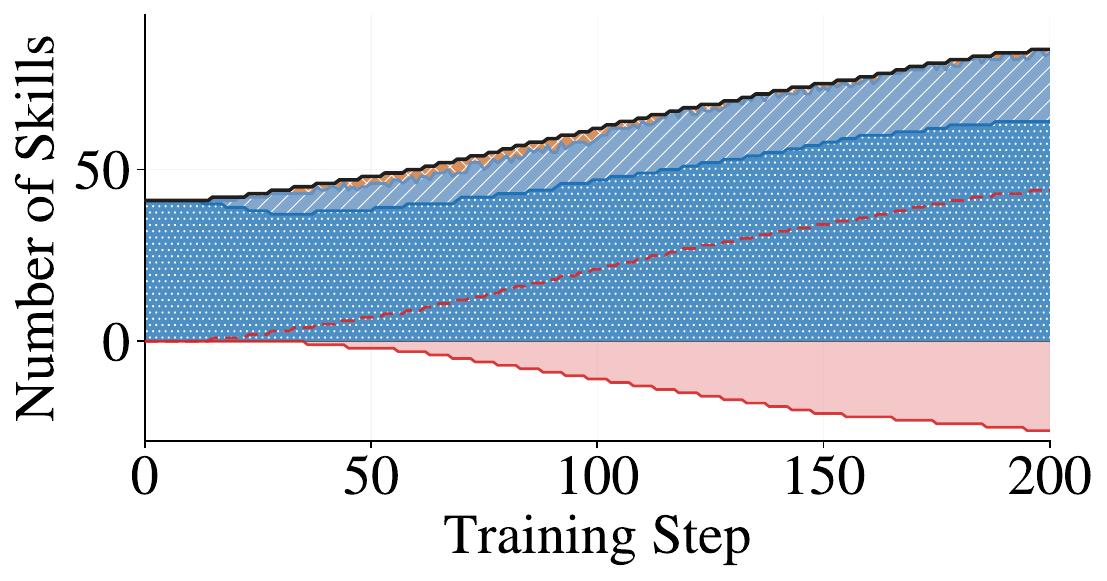}
        \caption{Search-Augmented QA}
    \end{subfigure}
    \caption{\textbf{Skill library dynamics} across three environments.
    \colorbox{clrStable}{\textcolor{white}{\small Stable}},
    \colorbox{clrActive}{\textcolor{white}{\small Active}},
    \colorbox{clrTrial}{\textcolor{white}{\small Trial}}: lifecycle states (stacked areas).
    \colorbox{clrRetired}{\textcolor{white}{\small Retired}}: cumulative retired (below zero).
    \textbf{---}: total non-retired;
    \textcolor{clrRetired}{\textbf{- -}}: mutated (gen$\geq$1).
    }
    \label{fig:forging}
\end{figure}

As shown in Figure~\ref{fig:forging}, the skill library evolves similarly across the three environments. On ALFWorld, 200 training steps grow the library to 100 non-retired skills, with 132 created and 32 retired; the cap binds near step 170, after which mutations and retirements balance into dynamic equilibrium. At the final checkpoint 76 skills are \texttt{stable} and 24 \texttt{active} or \texttt{trial}.

\begin{figure}[t]
    \centering
    \begin{subfigure}[t]{0.325\textwidth}
        \includegraphics[width=\textwidth]{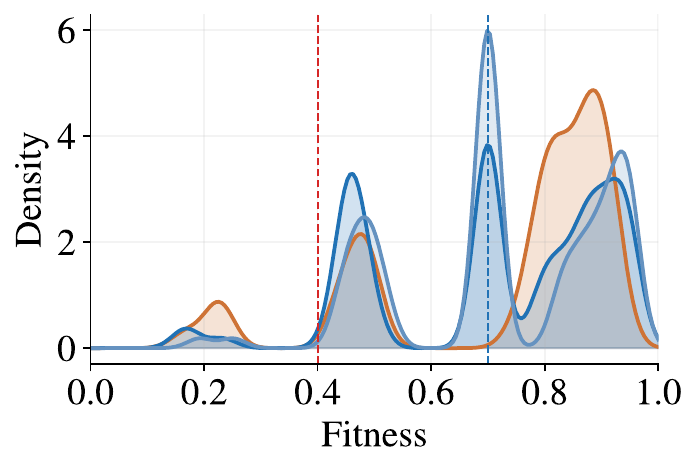}
        \caption{ALFWorld}
    \end{subfigure}
    \hfill
    \begin{subfigure}[t]{0.325\textwidth}
        \includegraphics[width=\textwidth]{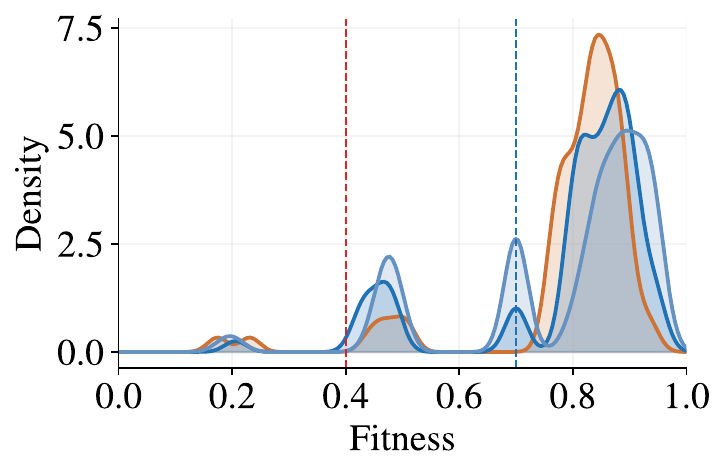}
        \caption{WebShop}
    \end{subfigure}
    \hfill
    \begin{subfigure}[t]{0.325\textwidth}
        \includegraphics[width=\textwidth]{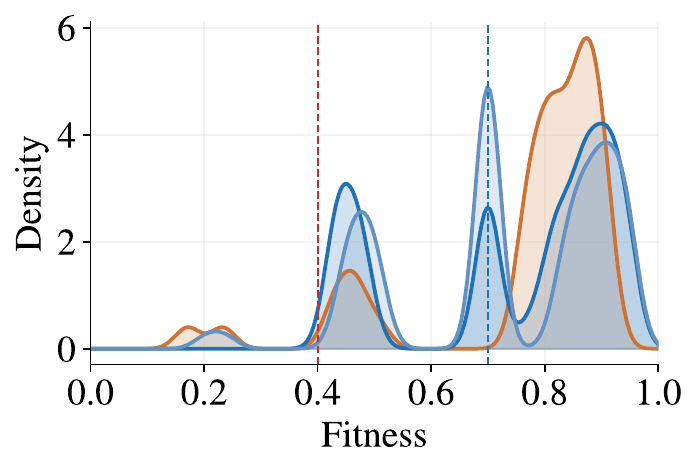}
        \caption{Search-Augmented QA}
    \end{subfigure}
    \caption{\textbf{Fitness distribution} at three training stages.
    \colorbox{clrTrial}{\textcolor{white}{\small Step 40}},
    \colorbox{clrStable}{\textcolor{white}{\small Step 120}},
    \colorbox{clrActive}{\textcolor{white}{\small Step 195}}: KDE at each stage.
    \textcolor{clrRetired}{\textbf{- -}}: $\delta_{\text{retire}}=0.4$;
    \textcolor{clrStable}{\textbf{- -}}: $\delta_{\text{stable}}=0.7$.
    Retirement depletes the left tail, mutation populates the middle band via newly born skills, and stabilization maintains the high-fitness regime.}
    \label{fig:fitness}
\end{figure}

As illustrated in Figure~\ref{fig:fitness}, the distribution evolves through two opposing flows rather than a monotone rightward shift. Retirement depletes the left tail: on ALFWorld the sub-$\delta_{\text{retire}}$ fraction shrinks from 7.5\% to 2.0\% across training, while mutation expands the middle band $[\delta_{\text{retire}},\delta_{\text{stable}})$ from 18.9\% to 22.0\% as children enter with a neutral prior at 0.5. The stable regime retains most of the library throughout and mean fitness stays near 0.73, an equilibrium in which pruning and churn rebalance. Appendix~\ref{app:cases} provides concrete mutation examples and the deepest forging lineages.

\section{Conclusion}
\label{sec:conclusion}

\method treats skill-library management as a fitness-driven forging process in which skills and the model co-evolve under online RL through retirement, stabilization, and LLM-guided mutation, delivering up to 7.8\% relative improvement over the strongest memory-augmented RL baseline across ALFWorld, WebShop, and Search-Augmented QA while keeping the active library compact. We release \skillfurnace, a 5k+ annotated record bundle of retirement-filtered SFT trajectories, evolved skill libraries with fitness annotations, and retirement events with human-annotated failure categories, in which procedural rigidity emerges as the dominant mode by which once-useful skills turn into active interference, the empirical signature of \textit{delayed obsolescence}.

\section{Limitations}
\label{sec:limitations}

Each configuration is trained with a single RL seed; the reported $\pm$ values aggregate five evaluation sampling seeds rather than independent training runs. Fitness is assigned at the episode level, so the $K$ co-retrieved skills share one binary reward; per-skill leave-one-out attribution agrees with this estimate on 93\% of retirement decisions, but the approximation may slow convergence where co-retrieval is structurally unavoidable. All experiments use a 7B base model, so lifecycle dynamics at larger scales remain open. Mutation relies on an external teacher; self-mutation by the base model still outperforms SkillRL but forgoes the mutation gain. \skillfurnace is drawn from three environments. More broadly, delayed obsolescence can affect any context accumulated for an improving agent, such as retrieved passages or tool catalogs, which a fitness-driven lifecycle could manage in the same way.

\bibliographystyle{unsrtnat}
\bibliography{main}

\newpage
\appendix

\section{Notation}
\label{app:notation}

Table~\ref{tab:notation} summarizes the symbols used throughout the paper, grouped by domain.

\begin{table*}[!h]
\caption{Notation used throughout the paper.}
\label{tab:notation}
\centering
\renewcommand{\arraystretch}{0.9}
\setlength{\tabcolsep}{6pt}
\resizebox{\textwidth}{!}{%
\begin{tabular}{l l l}
\toprule
\textbf{Symbol} & \textbf{Definition} & \textbf{Description} \\
\midrule
\multicolumn{3}{l}{\textit{Environment and Agent}} \\
\midrule
$d$ & Task description & Query input to the agent; $d_\tau$ denotes the instance tied to trajectory $\tau$ \\
$\mathcal{D}_{\text{task}}$ & Training task distribution & Uniform over per-environment training tasks \\
$R$ & Reward function & $R(\tau) \in \{0,1\}$, binary task success \\
$\pi_\theta$ & Agent policy & Parameterized by $\theta$ \\
$\pi_{\theta_0}$ & Base model policy & Before any fine-tuning \\
$\pi_{\theta_\text{old}}$ & Old policy (GRPO) & Generates rollouts; updated each training step \\
$\pi_{\theta_\text{sft}}$ & SFT checkpoint policy & RL initialization and reference policy $\pi_\text{ref}$ \\
$\tau = (p_1, a_1, o_2, \ldots)$ & Episode trajectory & Contexts, actions, observations over one episode \\
$a_t$ & Agent action at step $t$ & Sampled as $a_t \sim \pi_\theta(\cdot \mid p_t)$ \\
$o_t$ & Observation at step $t$ & Current environment observation \\
$p_t$ & Agent context at step $t$ & Built via Eq.~\eqref{eq:context}; $p^{(j)}$ uses library $\mathcal{S}^{(j)}$ at training step $j$ \\
$h_t$ & Interaction history & Sliding window of the last $W{=}5$ $(o,a)$ pairs \\
\midrule
\multicolumn{3}{l}{\textit{Skill Library}} \\
\midrule
$s_i = (v, b, \omega)$ & Skill & Name, behavioral principle, applicability condition \\
$\mathcal{S}$ & Skill library & Set of all non-retired skills \\
$\mathcal{S}_0$ & Initial skill library & Input to Eq.~\eqref{eq:proto_fitness}; $\mathcal{S}_{\text{retire}}^{(0)}$ is the pre-retired subset \\
$\mathcal{S}^{(k)}$ & Library at step $k$ & After $k$ forging updates; $\mathcal{S}^{(0)} = \mathcal{S}_0 \setminus \mathcal{S}_{\text{retire}}^{(0)}$ \\
$K$ & Retrieval count & Number of skills retrieved per episode \\
$\phi(d, \mathcal{S})$ & Retrieval function & Selects top-$K$ skills for task $d$; $\phi_{\text{eval}}$ is the pre-retirement variant \\
$\mathcal{S}_\tau$ & Retrieved skill set & $\phi(d, \mathcal{S})$ for trajectory $\tau$ \\
$U_i$ & Usage count & Cumulative retrievals of $s_i$; superscript $(0)$ marks pre-retirement value \\
$C_i$ & Success count & Cumulative episodes in which $s_i$ was retrieved and succeeded \\
$f_i$ & Skill fitness & $C_i/U_i$ when $U_i \geq N_{\text{warm}}$, else $f_{\text{default}}$ \\
$f_{\text{default}}$ & Default fitness & Neutral prior for skills with $U_i < N_{\text{warm}}$ \\
$\hat{f}_i$ & Proto-fitness & $C_i^{(0)}/U_i^{(0)}$ from pre-retirement rollouts; equals $f_i^{(0)}$ \\
$f_i^{(k)}$ & Fitness at step $k$ & Non-stationary estimate (Eq.~\eqref{eq:nonstationary}) \\
$\ell(s_i)$ & Lifecycle state & $\in \{\texttt{trial}, \texttt{active}, \texttt{stable}, \texttt{retired}\}$ \\
$g_i$ & Generation & Mutation depth; $g_i=0$ for distilled skills \\
$\mathcal{M}_{\text{cand}}$ & Mutation candidate pool & Active skills with moderate fitness (Eq.~\eqref{eq:mutation}) \\
$\mathcal{M}_T$ & Teacher model & Generates mutations via $\mathcal{P}_\mu$ \\
$\mathcal{P}_\mu$ & Mutation template & Assembles parent skill, fitness, and failures \\
$\mathcal{T}_{\text{fail}}$ & Failure trajectories & Recent failed episodes used as mutation context \\
$\mathcal{D}_{\text{SFT}}$ & SFT dataset & Successful trajectories filtered by retirement (Eq.~\eqref{eq:sft_data}) \\
$\operatorname{Forge}(\cdot)$ & Forging operator & Full forging cycle (Algorithm~\ref{alg:forging}) \\
\midrule
\multicolumn{3}{l}{\textit{Hyperparameters}} \\
\midrule
$M$ & Preliminary eval episodes & Rollout count for pre-retirement \\
$\delta_{\text{pre}}$ & Pre-retirement threshold & Proto-fitness below which initial skills are discarded \\
$\delta_{\text{retire}}$ & Retirement threshold & Fitness boundary below which active skills are retired \\
$\delta_{\text{stable}}$ & Stabilization threshold & Fitness boundary for active $\to$ stable \\
$\delta_{\text{demote}}$ & Demotion threshold & Fitness boundary for stable $\to$ active \\
$[\delta_{\text{low}}, \delta_{\text{high}}]$ & Mutation fitness range & Bounds for mutation eligibility \\
$N_{\text{min}}$ & Proto-fitness min usage & Min usage before proto-fitness is evaluated \\
$N_{\text{warm}}$ & Fitness warm-up & Min usage before fitness replaces $f_{\text{default}}$ \\
$N_{\text{promote}}$ & Promotion min usage & Min usage for trial $\to$ active \\
$N_{\text{stable}}$ & Stabilization min usage & Min usage for active $\to$ stable \\
$N_{\text{mutate}}$ & Mutation min usage & Min usage for mutation eligibility \\
$N_{\text{retire}}$ & Retirement min usage & Min usage before an active skill can be retired \\
$N_{\text{retire}}^{(g_i)}$ & Generation-aware retirement & $N_{\text{protect}}$ if $g_i=0$, else $N_{\text{retire}}$ \\
$N_{\text{protect}}$ & Protection threshold & Higher usage bar for generation-0 skills \\
$K_{\text{retire}}$ & Retirement budget & Max retirements per forging cycle \\
$K_{\text{mutate}}$ & Mutation budget & Max mutations per forging cycle \\
$S_{\text{max}}$ & Skill cap & Maximum library size \\
$E$ & Forging period & Validation epochs between forging cycles \\
$W$ & History window size & Recent $(o,a)$ pairs kept in $h_t$; $W{=}5$ \\
\midrule
\multicolumn{3}{l}{\textit{GRPO Training}} \\
\midrule
$\mathcal{L}_{\text{SFT}}$ & SFT loss & Cross-entropy on retirement-filtered trajectories (Eq.~\eqref{eq:sft_loss}) \\
$G$ & Group size & Trajectories sampled per task \\
$\hat{A}_n$ & Group-normalized advantage & $(r_n - \mu_G) / \sigma_G$ \\
$\epsilon$ & Clipping range & PPO-style surrogate clipping \\
$\beta$ & KL coefficient & Regularization against reference policy \\
$\tau_n$ & Episode $n$ in a GRPO group & Used for $\hat{A}_n$ and reward $r_n = R(\tau_n)$ \\
$\boldsymbol{y}_n$ & Token sequence of $\tau_n$ & Action tokens produced autoregressively by $\pi_\theta$ \\
$\tau^{(j)}$ & Trajectory at step $j$ & Rollout from $\pi_{\theta_j}$ for the non-stationary fitness (Eq.~\eqref{eq:nonstationary}) \\
$y_{n,l}$ & Token at position $l$ & $l$-th action token in $\boldsymbol{y}_n$ \\
$c_{n,l}$ & Left context of $y_{n,l}$ & Environment-step context plus earlier tokens \\
$\rho_{n,l}$ & Importance ratio & $\pi_\theta(y_{n,l}\mid c_{n,l}) / \pi_{\theta_\text{old}}(y_{n,l}\mid c_{n,l})$ \\
\bottomrule
\end{tabular}
}
\end{table*}

\section{Implementation Details}
\label{app:hyperparams}

\paragraph{Training configuration.}
Table~\ref{tab:training} lists the training hyperparameters and hardware setup.
The base model is Qwen2.5-7B-Instruct~\citep{qwen}, fine-tuned with GRPO on 64$\times$ NVIDIA H200 GPUs using the verl framework.
The initial skill library is environment-specific: ALFWorld uses 44 skills (12 general, 21 task-specific, 11 common-mistake), WebShop uses 66 skills (15 general, 39 task-specific, 12 common-mistake), and Search-Augmented QA uses 41 skills (10 general, 20 query-type, 11 common-mistake), all inherited from the SkillRL release~\citep{skillrl}. General skills encode cross-task heuristics (e.g., ``always verify inventory after pickup''), task-specific skills target individual task types, and common-mistake skills capture frequently observed failure patterns from initial rollouts (e.g., ``do not leave an object behind after finding it'').

\paragraph{Evaluation sample sizes.}
The three environments follow the setup of SkillRL~\citep{skillrl}, our most direct baseline, and cover three distinct task types: embodied control, web navigation, and retrieval-based QA. ALFWorld reports per-sub-task accuracy on the 130-task unseen-test set, partitioned as Pick (24), Look (15), Clean (27), Heat (26), Cool (21), and Pick2 (17); the \textbf{All} column in Table~\ref{tab:main1} reports overall success on this set, and baseline rows are taken from SkillRL~\citep{skillrl}. WebShop uses the standard 500-task held-out test set for both Score and Succ. Search-Augmented QA per-sub-dataset sample sizes appear in Table~\ref{tab:searchqa_full}.

\begin{table}[h]
\caption{Training configuration for \method. All three environments share the GRPO optimizer, group size, and training budget; hardware is identical across runs.}
\label{tab:training}
\centering
\small
\setlength{\aboverulesep}{2pt}
\setlength{\belowrulesep}{2pt}
\setlength{\tabcolsep}{30pt}
\begin{tabular}{lc}
\toprule
\textbf{Hyperparameter} & \textbf{Value} \\
\midrule
Base model & Qwen2.5-7B-Instruct \\
GRPO learning rate & $1 \times 10^{-6}$ \\
Batch size & 16 \\
Group size $G$ & 8 \\
Clipping range $\epsilon$ & 0.2 \\
KL coefficient $\beta$ & 0.001 \\
Training steps & 200 \\
History window size $W$ & 5 \\
Sampling temperature (train \& eval) & 1.0 \\
Validation epoch & 5 steps \\
Hardware & 64$\times$ NVIDIA H200 \\
Framework & verl \\
\bottomrule
\end{tabular}
\end{table}

\paragraph{Forging hyperparameters.}
Table~\ref{tab:hyperparams} lists all forging-specific hyperparameters. All thresholds and budgets are shared across the three environments rather than tuned per task; only $M$ and $S_{\text{max}}$ scale with the size of the seed library. The pre-retirement threshold $\delta_{\text{pre}}=0.3$ is looser than the runtime $\delta_{\text{retire}}=0.4$ because the pre-retirement phase uses the unadapted base policy $\pi_{\theta_0}$, whose success rates are systematically lower than those observed during RL; a lower bar avoids discarding borderline-but-potentially-useful skills before RL has had a chance to re-evaluate them. Two usage-count thresholds also deserve clarification. The proto-fitness minimum $N_{\text{min}}=3$ (Eq.~\eqref{eq:proto_fitness}) is smaller than the runtime warm-up $N_{\text{warm}}=5$ (Eq.~\eqref{eq:fitness}) because proto-fitness is estimated under the fixed base policy over $M$ dedicated rollouts, so three observations already yield a low-variance estimate for the binary filter $\hat{f}_i < \delta_{\text{pre}}$. Runtime fitness, in contrast, is estimated under a moving policy and feeds into continuous lifecycle transitions (promotion, stabilization, demotion, retirement), which justifies a slightly more conservative warm-up. Correspondingly, Eq.~\eqref{eq:proto_fitness} is a hard \textit{filter}: skills without $U_i^{(0)} \geq N_{\text{min}}$ are simply not candidates for removal and remain in $\mathcal{S}^{(0)}$. In contrast, Eq.~\eqref{eq:fitness} must emit a numeric score for retrieval and ranking, so it falls back to the neutral prior $f_{\text{default}} = 0.5$ when the usage count is too low to trust the empirical rate. A second boundary case arises because $\delta_{\text{low}}=\delta_{\text{retire}}=0.4$: a skill with $f_i$ exactly equal to this value satisfies both the retirement condition (strict inequality $f_i < \delta_{\text{retire}}$ is false, so retirement does not fire) and the mutation range (closed interval $[\delta_{\text{low}}, \delta_{\text{high}}]$ includes the endpoint), so such a skill is eligible for mutation but not retirement and behaves identically to skills with $f_i$ slightly above $0.4$.

\begin{table}[h]
\caption{Forging hyperparameters for \method. Slash-separated values apply to ALFWorld, WebShop, and Search-Augmented QA respectively; $S_{\text{max}}$ and $M$ scale with the initial library size of each environment.}
\label{tab:hyperparams}
\centering
\small
\setlength{\aboverulesep}{2pt}
\setlength{\belowrulesep}{2pt}
\resizebox{\columnwidth}{!}{%
\setlength{\tabcolsep}{20pt}
\begin{tabular}{lcc}
\toprule
\textbf{Hyperparameter} & \textbf{Symbol} & \textbf{Value} \\
\midrule
Proto-fitness min usage & $N_{\text{min}}$ & 3 \\
Pre-retirement threshold & $\delta_{\text{pre}}$ & 0.3 \\
Preliminary eval episodes & $M$ & 500 / 450 / 400 \\
Fitness warm-up threshold & $N_{\text{warm}}$ & 5 \\
Default fitness (new skills) & $f_{\text{default}}$ & 0.5 \\
Promotion min usage & $N_{\text{promote}}$ & 10 \\
Retirement fitness threshold & $\delta_{\text{retire}}$ & 0.4 \\
Retirement min usage & $N_{\text{retire}}$ & 20 \\
Stabilization fitness threshold & $\delta_{\text{stable}}$ & 0.7 \\
Stabilization min usage & $N_{\text{stable}}$ & 30 \\
Demotion fitness threshold & $\delta_{\text{demote}}$ & 0.5 \\
Mutation fitness range & $[\delta_{\text{low}}, \delta_{\text{high}}]$ & $[0.4, 0.7]$ \\
Mutation min usage & $N_{\text{mutate}}$ & 5 \\
Max mutations per cycle & $K_{\text{mutate}}$ & 5 \\
Max retirements per cycle & $K_{\text{retire}}$ & 3 \\
Total skill cap & $S_{\text{max}}$ & 100 / 95 / 85 \\
Original skill protection & $N_{\text{protect}}$ & 50 \\
Forging frequency & $E$ & 2 (every 10 training steps) \\
Teacher model & $\mathcal{M}_T$ & Kimi-K2.5~\citep{kimik25} \\
Skill retrieval & top-$K$ & 6 (task-type matching) \\
\bottomrule
\end{tabular}
}
\end{table}

\paragraph{Lifecycle well-definedness.}
The four branches of Eq.~\eqref{eq:transitions} have pairwise disjoint guards: they require $\ell(s_i) \in \{\texttt{trial}, \texttt{active}, \texttt{stable}, \texttt{active}\}$ respectively, and the two \texttt{active} branches trigger on disjoint fitness ranges, one requiring $f_i \geq \delta_{\text{stable}}$ and the other $f_i < \delta_{\text{retire}}$, so at most one branch fires per skill per cycle. Eq.~\eqref{eq:transitions} is declarative; Algorithm~\ref{alg:forging} specifies the per-cycle ordering, with promotion and demotion preceding retirement and stabilization preceding mutation.

\paragraph{Mutation weighting.}
We use the simplest monotone-decreasing weighting, $P(s_i) \propto 1-f_i$, rather than sharper alternatives such as $1/f_i$ or $\exp(-\lambda f_i)$. Because mutation candidates already satisfy $\delta_{\text{low}} \leq f_i \leq \delta_{\text{high}}$ (default $[0.4, 0.7]$), the weights $1 - f_i$ lie in $[0.3, 0.6]$, giving at most a $2\times$ preference for lower-fitness parents. This is enough to bias selection toward skills with more room to improve without concentrating mutation on a single outlier. Empirically, both $1/f_i$ and uniform sampling lower average success (Table~\ref{tab:design_alternatives}).

\paragraph{Threshold sensitivity.}
Table~\ref{tab:sensitivity} sweeps the two most load-bearing thresholds, the retirement threshold $\delta_{\text{retire}}$ and the library cap $S_{\text{max}}$, around their default values on ALFWorld. The overall success rate varies within $\pm 1.4\%$ across the swept range, showing robustness to threshold choices. We observe a mild concavity: overly aggressive retirement ($\delta_{\text{retire}}=0.5$) removes borderline-but-useful skills before they stabilize, while overly permissive retirement ($\delta_{\text{retire}}=0.3$) retains low-fitness skills that dilute the retrieved context. The default $\delta_{\text{retire}}=0.4$ sits near the peak of this trade-off.

\begin{table}[h]
\caption{Sensitivity of overall ALFWorld success (\%) to the two most load-bearing thresholds, sweeping one threshold at a time around its default. Default values are boxed.}
\label{tab:sensitivity}
\centering
\small
\setlength{\aboverulesep}{2pt}
\setlength{\belowrulesep}{2pt}
\setlength{\tabcolsep}{14pt}
\begin{tabular}{lcccc}
\toprule
\multicolumn{5}{l}{\textit{Retirement threshold $\delta_{\text{retire}}$ (default 0.4)}} \\
$\delta_{\text{retire}}$ & 0.3 & \fbox{0.4} & 0.5 & 0.6 \\
Success (\%) & 91.3 & \best{92.4} & 91.8 & 91.0 \\
\midrule
\multicolumn{5}{l}{\textit{Library cap $S_{\text{max}}$ (default 100)}} \\
$S_{\text{max}}$ & 70 & 85 & \fbox{100} & 130 \\
Success (\%) & 91.0 & 91.9 & \best{92.4} & 91.6 \\
\bottomrule
\end{tabular}
\end{table}

\paragraph{Compute resources.}
All RL runs use 64$\times$ NVIDIA H200 GPUs under the verl framework. Estimated wall-clock per 200-step training run is ${\approx}24$ hours for ALFWorld, ${\approx}18$ hours for WebShop, and ${\approx}72$ hours for Search-Augmented QA (retrieval-dominated), for a total of ${\approx}114$ wall-clock hours $\times$ 64 GPUs $= 7{,}296$ GPU-hours across the three environments per full training run. The pre-retirement evaluation phase (Section~\ref{sec:presft}) adds ${\approx}6$ hours per environment on the same hardware. Evaluation is run five times per reported number (sampling seeds $\{30, 40, 42, 50, 62\}$) and adds ${\approx}1.5$--$4.5$ hours per environment per seed. Teacher model overhead is negligible relative to RL: a forging cycle runs every $E=2$ validation epochs (every 10 training steps) and issues up to $K_{\text{mutate}}=5$ mutation calls per cycle, for at most $\lceil 200/10 \rceil \times 5 = 100$ mutation calls per training run, all dispatched asynchronously during validation so they do not block GPU training. The teacher $\mathcal{M}_T$ is treated as a frozen black-box LLM: any sufficiently capable instruction-tuned model (proprietary or open-weight) can serve in this role, since the interface is only a structured template $\mathcal{P}_\mu$ and a single-shot text response. We use Kimi-K2.5 here for mutation quality, but the framework imposes no dependence on a specific teacher: GLM-5.2~\citep{glm52} and MiniMax M3~\citep{minimaxm3} perform on par with it (Table~\ref{tab:teacher_capability}).

\paragraph{Mutation template.}
\label{app:mutation_prompt}
The mutation operator (Section~\ref{sec:lifecycle}, Eq.~\eqref{eq:mutation}) assembles the parent skill's metadata, fitness statistics, and recent failure examples into the following structured template $\mathcal{P}_\mu$, which is sent to the teacher model $\mathcal{M}_T$:

\begin{tcolorbox}[colback=gray!5, colframe=gray!60, fontupper=\small\ttfamily, title={\small\sffamily Mutation Template $\mathcal{P}_\mu$}, breakable]
You are improving a skill for an embodied AI agent in \{environment\}.\\[4pt]
The following skill has moderate performance (fitness=\{$f_i$\}, used \{$U_i$\} times, succeeded \{$C_i$\} times):\\[4pt]
Original skill:\\
- ID: \{skill\_id\}\\
- Title: \{title\}\\
- Principle: \{principle\}\\
- When to apply: \{when\_to\_apply\}\\[4pt]
\textit{[If recent failures exist:]}\\
This skill was involved in the following failed episodes:\\
- Task: \{task description\}\\
\hspace*{4pt}Agent action: \{failed action\}\\
- ...\\
Please analyze these failure patterns and generate an improved skill that specifically addresses them.\\[4pt]
Generate ONE improved variant that addresses the weakness shown by its moderate success rate.
The improved skill should be more specific, actionable, or correct than the original.\\[4pt]
Return ONLY a JSON object with these fields:\\
\{"skill\_id": "\{new\_id\}", "title": "...", "principle": "...", "when\_to\_apply": "..."\}
\end{tcolorbox}

The template includes up to 8 recent failure examples (task description and agent action, each truncated to 200 characters) from episodes where the parent skill was retrieved but the episode failed. Failure examples are omitted when no failures are recorded. The teacher model returns a single JSON object, which is parsed and assigned a new skill identifier.

\section{\skillfurnace Dataset Details}
\label{app:skillfurnace}

\skillfurnace is a three-part dataset released alongside \method. Each part draws from a distinct data source within our experimental pipeline: (i)~\textbf{SFT trajectories} come from rollouts of the Qwen2.5-7B-Instruct base model during the preliminary evaluation phase, before any RL training has begun; (ii)~\textbf{evolved skill libraries} are initialized from the seed library released by SkillRL~\citep{skillrl} and then logged at every forging cycle as one record per skill per snapshot; (iii)~\textbf{retirement events} are lifecycle-gate decisions aggregated across the primary training run and additional previously recorded training runs (ablation and sensitivity variants), so that the released event count is not limited to a single run. Every mined event is labelled and quality-controlled under the same criteria as the primary run. In total, \skillfurnace contains \textbf{5{,}852 annotated records} across the three environments: 1{,}013 SFT trajectories, 4{,}521 library snapshot records, and 318 retirement events, summarized in Table~\ref{tab:skillfurnace}. Table~\ref{tab:skillfurnace_library} summarizes the library component across environments, Table~\ref{tab:skillfurnace} reports dataset scale and retirement-event statistics, and Table~\ref{tab:failure_taxonomy} breaks retirement events down by failure category per environment.

\begin{table}[h]
\caption{\skillfurnace evolved-library composition per environment. Initial $|\mathcal{S}_0|$ is the seed library released by SkillRL~\citep{skillrl}; \emph{total ever created} counts every skill ever introduced, including all mutation children; \emph{retired during training} counts skills that reached \texttt{retired}; \emph{final non-retired} is the active library size at the last training step. The final row shows what fraction of the surviving library originated from LLM-guided mutation.}
\label{tab:skillfurnace_library}
\centering
\small
\setlength{\aboverulesep}{2pt}
\setlength{\belowrulesep}{2pt}
\resizebox{\columnwidth}{!}{%
\setlength{\tabcolsep}{18pt}
\begin{tabular}{lcccc}
\toprule
\textbf{Statistic} & \textbf{ALFWorld} & \textbf{WebShop} & \textbf{Search-Aug. QA} & \textbf{Total} \\
\midrule
Initial $|\mathcal{S}_0|$ & 44 & 66 & 41 & 151 \\
Total ever created & 132 & 121 & 111 & 364 \\
Retired during training & 32 & 26 & 26 & 84 \\
Final non-retired & 100 & 95 & 85 & 280 \\
\quad of which \texttt{stable} & 76 & 80 & 64 & 220 \\
\quad of which mutated ($g_i \geq 1$) & 57 & 29 & 44 & 130 \\
\bottomrule
\end{tabular}
}
\end{table}

\paragraph{SFT trajectories.}
The preliminary evaluation phase executes $M$ rollout episodes with the Qwen2.5-7B-Instruct base model $\pi_{\theta_0}$ and retrieved skills injected via Eq.~\eqref{eq:context}. Per-environment $M$ values are 500 (ALFWorld), 450 (WebShop), and 400 (Search-Augmented QA); see Table~\ref{tab:hyperparams} for all forging-related hyperparameters. We retain every successful trajectory whose retrieved-skill set is disjoint from the pre-retired set $\mathcal{S}_{\text{retire}}^{(0)}$, yielding the SFT corpus $\mathcal{D}_{\text{SFT}}$ defined in Eq.~\eqref{eq:sft_data}: 412 trajectories in ALFWorld, 348 in WebShop, and 253 in Search-Augmented QA, for a total of 1{,}013 released SFT trajectories. Each trajectory is stored as an ordered sequence of context--action pairs with the retrieved-skill ids attached to each context, so downstream practitioners can reconstruct the base corpus and re-filter it under alternative retention rules without rerunning rollouts. All trajectories in this component come exclusively from the base Qwen2.5 model, making them a clean cold-start fine-tuning resource decoupled from any specific RL setup.

\paragraph{Evolved skill libraries.}
Every forging cycle emits a library snapshot recording, for each skill, the id, name, behavioral principle $b$, applicability condition $\omega$, cumulative usage count $U_i$, cumulative success count $C_i$, current fitness $f_i$, lifecycle state $\ell(s_i) \in \{\texttt{trial}, \texttt{active}, \texttt{stable}, \texttt{retired}\}$, generation depth $g_i$, and parent id for mutated skills. The initial library $\mathcal{S}_0$ logged at snapshot 0 is the seed library released by SkillRL~\citep{skillrl}; subsequent snapshots track how this seed evolves under RL. Forging runs every 10 training steps over the 200-step budget, producing 21 snapshots per run. Each snapshot emits one record per skill tracked at that step, yielding 1{,}512 records in ALFWorld, 1{,}683 in WebShop, and 1{,}326 in Search-Augmented QA, for a total of 4{,}521 library snapshot records. Table~\ref{tab:skillfurnace_library} reports the composition at the last training step. Across the three environments, the library grows from 44, 66, and 41 seed skills to a final non-retired size of 100, 95, and 85 at the cap $S_{\text{max}}$, with 132, 121, and 111 skills created cumulatively. Of the final non-retired skills, 76, 80, and 64 reach the \texttt{stable} state, and 57, 29, and 44 originate from LLM-guided mutation ($g_i \geq 1$), making 46.4\% of final skills across the three environments children rather than seed skills. Snapshots are released together with per-cycle forging logs, so any fitness trajectory or lifecycle transition can be reconstructed offline.

\begin{figure}[t]
    \centering
    \begin{subfigure}[t]{0.32\textwidth}
        \includegraphics[width=\textwidth]{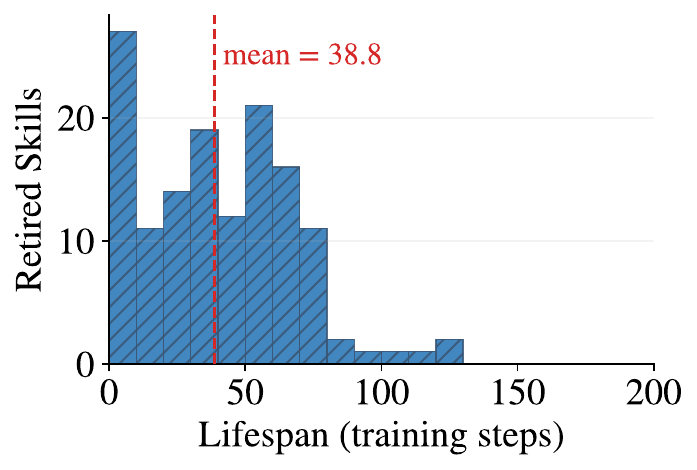}
        \caption{ALFWorld}
    \end{subfigure}
    \hfill
    \begin{subfigure}[t]{0.32\textwidth}
        \includegraphics[width=\textwidth]{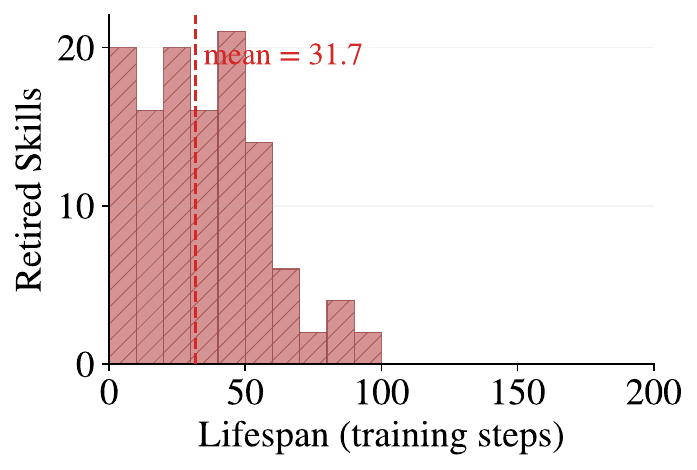}
        \caption{WebShop}
    \end{subfigure}
    \hfill
    \begin{subfigure}[t]{0.32\textwidth}
        \includegraphics[width=\textwidth]{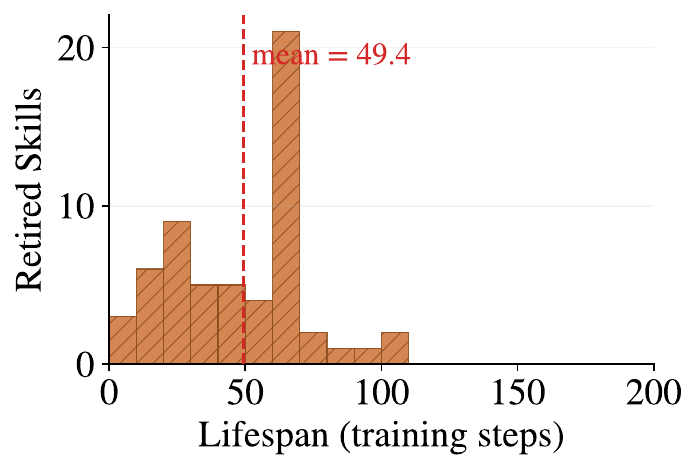}
        \caption{Search-Augmented QA}
    \end{subfigure}
    \caption{\textbf{Lifespan distribution of retirement events in \skillfurnace}, measured as $\texttt{retired\_step} - \texttt{born\_step}$. Retirements spread across the full training horizon rather than clustering at any single stage; the red dashed line marks the per-environment mean.}
    \label{fig:lifespan}
\end{figure}

\paragraph{Retirement events.}
Each retirement event, whether triggered by the retirement gate or by the library cap $S_{\text{max}}$, records the full autopsy of a removed skill: the skill content, the complete fitness history leading up to retirement, the lifecycle-transition log, parent and sibling skills from the mutation tree, a human-annotated failure category drawn from the 8-way taxonomy described in Appendix~\ref{app:taxonomy}, and a free-text \textit{key flaw} field describing the retired skill's failure mode. To make the released sample size independent of any single training configuration, events are aggregated from the primary training run and additional previously recorded training runs; the per-environment totals in Table~\ref{tab:skillfurnace} therefore exceed the primary-run retirement counts reported in Section~\ref{sec:evo_dynamics} by design. Retirement events exhibit an average peak fitness of 0.67 and an average lifespan of 38.0 training steps, with fitness dropping by 0.28 on average at retirement; 57.2\% first reached the \texttt{stable} state before this decay, the defining pattern of \textit{delayed obsolescence} that motivates the forging lifecycle. Figure~\ref{fig:lifespan} shows the lifespan distribution: retirements spread across the full training horizon rather than clustering at any single stage, so the retirement gate fires throughout RL rather than only at initialization or endgame. Table~\ref{tab:failure_taxonomy} reports the per-environment breakdown across the eight categories. Overly rigid ordering is the dominant mode in every environment (74.6\% / 77.7\% / 67.8\%), confirming that procedural rigidity rather than content omission is the principal failure mode of LLM-distilled skills. Environment-specific patterns also emerge: \textit{hallucinated action} accounts for 10.7\% of WebShop events, corresponding to skills that reference UI elements absent from the WebShop environment, and \textit{contradicts environment} is exclusive to Search-Augmented QA, corresponding to skills that invoke search operations the environment does not expose.

\begin{table}[h]
\caption{\skillfurnace retirement events by failure category (8-way human-annotated taxonomy). Percentages are within environment. \textit{Overly rigid ordering} dominates in every environment, confirming that procedural rigidity, not content omission, is the principal failure mode of LLM-distilled skills. Environment-specific patterns also emerge: \textit{hallucinated action} appears only in WebShop (UI elements absent from the environment) and \textit{contradicts environment} appears only in Search-Augmented QA (search operations absent from the environment).}
\label{tab:failure_taxonomy}
\centering
\small
\setlength{\aboverulesep}{2pt}
\setlength{\belowrulesep}{2pt}
\resizebox{\columnwidth}{!}{%
\setlength{\tabcolsep}{12pt}
\begin{tabular}{lcccc}
\toprule
\textbf{Failure category} & \textbf{ALFWorld} & \textbf{WebShop} & \textbf{Search-Aug. QA} & \textbf{Total} \\
\midrule
Overly rigid ordering     & 103 (74.6\%) & 94 (77.7\%) & 40 (67.8\%) & 237 (74.5\%) \\
Overly specific           & 26 (18.8\%)  & 2  (1.7\%)  & 2  (3.4\%)  & 30  (9.4\%) \\
Premature specialization  & 6  (4.3\%)   & 10 (8.3\%)  & 0  (0.0\%)  & 16  (5.0\%) \\
Hallucinated action       & 0  (0.0\%)   & 13 (10.7\%) & 0  (0.0\%)  & 13  (4.1\%) \\
Redundant with policy     & 2  (1.4\%)   & 1  (0.8\%)  & 7  (11.9\%) & 10  (3.1\%) \\
Contradicts environment   & 0  (0.0\%)   & 0  (0.0\%)  & 7  (11.9\%) & 7   (2.2\%) \\
Scope mismatch            & 0  (0.0\%)   & 0  (0.0\%)  & 3  (5.1\%)  & 3   (0.9\%) \\
Vague or tautological     & 1  (0.7\%)   & 1  (0.8\%)  & 0  (0.0\%)  & 2   (0.6\%) \\
\midrule
\textbf{Total}            & \textbf{138} & \textbf{121} & \textbf{59} & \textbf{318} \\
\bottomrule
\end{tabular}
}
\end{table}

\paragraph{Release format.}
\skillfurnace is released as three top-level JSON files (\texttt{sft\_trajectories.json}, \texttt{library\_snapshots.json}, \texttt{retirement\_events.json}) with a schema document. All records are keyed by a global \texttt{skill\_id} so the three components can be joined: every skill id appearing in \texttt{retirement\_events.json} is also present in the matching run's \texttt{library\_snapshots.json} for every step between birth and retirement, and every skill retrieved in \texttt{sft\_trajectories.json} is resolvable in the step-0 library snapshot. The dataset is released under CC-BY-4.0.

\section{Search-Augmented QA Full Per-Dataset Results}
\label{app:searchqa_table}

Table~\ref{tab:searchqa_full} reports accuracy on each of the seven Search-Augmented QA sub-datasets that are summarised by the radar chart in Figure~\ref{fig:searchqa_radar}. Both in-domain datasets ($\dagger$: NQ, HotpotQA) and out-of-domain datasets ($\star$: TriviaQA, PopQA, 2WikiMultiHopQA, MuSiQue, Bamboogle) are included. The 48.7\% aggregate number reported in the main text corresponds to micro-averaged accuracy over the 51{,}713-sample concatenation of these seven test sets.

\begin{table}[h]
\caption{Per-dataset accuracy on Search-Augmented QA (\%). $^\dagger$: in-domain; $^\star$: out-of-domain. Per-dataset cells for non-\method rows are taken from their respective source papers. The \textbf{Overall} column is recomputed as the sample-weighted micro-average over the 51{,}713-sample union shown in \textit{Test samples}, so that all methods are compared on an identical test set; these values may differ from source-paper totals that used different evaluation configurations. \best{Bold}: best; \second{underline}: second best. ``--'': not reported.}
\label{tab:searchqa_full}
\centering
\small
\setlength{\aboverulesep}{2pt}
\setlength{\belowrulesep}{2pt}
\resizebox{\textwidth}{!}{%
\setlength{\tabcolsep}{6pt}
\begin{tabular}{l ccccccc c}
\toprule
\textbf{Method} & \textbf{NQ}$^\dagger$ & \textbf{TriviaQA}$^\star$ & \textbf{PopQA}$^\star$ & \textbf{HotpotQA}$^\dagger$ & \textbf{2Wiki}$^\star$ & \textbf{MuSiQue}$^\star$ & \textbf{Bamboogle}$^\star$ & \textbf{Overall} \\
\textit{Test samples} & \textit{3{,}610} & \textit{11{,}313} & \textit{14{,}267} & \textit{7{,}405} & \textit{12{,}576} & \textit{2{,}417} & \textit{125} & \textit{51{,}713} \\
\midrule
\multicolumn{9}{l}{\textit{In-context / retrieval baselines}} \\
Vanilla~\citep{qwen}            & 11.6 & 35.6 & 1.2  & 16.4 & 22.2 & 4.8  & 14.4 & 16.9 \\
CoT                             & 12.8 & 35.6 & 3.8  & 16.2 & 22.6 & 6.6  & 24.0 & 17.9 \\
RAG                             & 27.4 & 58.2 & 17.8 & 25.8 & 23.2 & 9.4  & 16.8 & 29.4 \\
Search-o1~\citep{searcho1}      & 19.4 & 40.6 & 11.4 & 17.0 & 27.0 & 8.6  & 30.4 & 22.9 \\
R1-Instruct~\citep{deepseekr1}  & 21.0 & 44.9 & 17.1 & 20.8 & 27.5 & 6.0  & 19.2 & 26.0 \\
\midrule
\multicolumn{9}{l}{\textit{RL-based search methods}} \\
Search-R1~\citep{searchr1}      & 39.3 & 61.0 & 39.7 & 37.0 & 40.1 & 14.6 & 36.8 & 42.9 \\
ZeroSearch~\citep{zerosearch}   & 43.6 & 61.8 & \best{51.5} & 34.6 & 35.2 & 18.4 & 27.8 & 45.2 \\
StepSearch~\citep{stepsearch}   & --   & --   & --   & 38.6 & 36.6 & \best{22.6} & 40.0 & -- \\
EvolveR~\citep{evolver}         & 43.5 & \second{63.4} & 44.6 & 38.2 & \best{42.0} & 15.6 & 54.4 & 45.8 \\
SkillRL~\citep{skillrl}         & \second{45.9} & 63.3 & 45.9 & \second{43.2} & 40.3 & 20.2 & \second{73.8} & \second{46.8} \\
\ours\best{\method (Ours)}      & \ours\best{48.2} & \ours\best{65.0} & \ours\second{50.1} & \ours\best{43.9} & \ours\second{40.4} & \ours\second{20.3} & \ours\best{77.2} & \ours\best{48.7} \\
\bottomrule
\end{tabular}
}
\end{table}

\method obtains the best or second-best accuracy on every sub-dataset, with the largest absolute improvements on the multi-hop benchmarks Bamboogle ($+3.4$ over the strongest baseline) and HotpotQA ($+0.7$). Gains are consistent on both in-domain and out-of-domain splits, indicating that the forging lifecycle does not overfit to the training distribution.

\section{Case Studies}
\label{app:cases}

\subsection{Deepest Forging Lineages}
\label{app:lineages}

\begin{figure}[h]
    \centering
    \includegraphics[width=\textwidth]{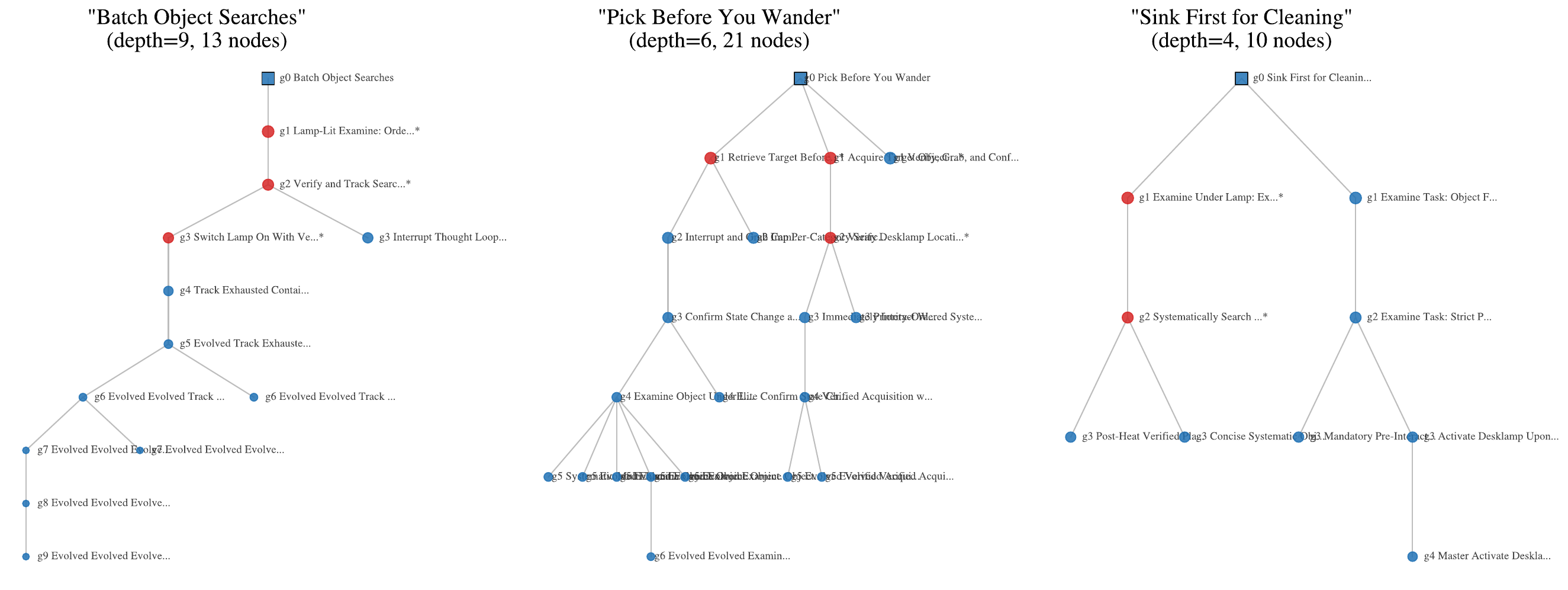}
    \caption{\textbf{Deepest forging lineages} on ALFWorld. Each subplot shows a root skill (gen=0, square) and its multi-generation descendants (circles). Red = retired, blue = stable. Early variants are often retired while later, more refined variants survive.}
    \label{fig:lineages}
\end{figure}

As shown in Figure~\ref{fig:lineages}, mutation lineages exhibit a heavy-tailed depth distribution: most refinements converge to a stable variant within one or two generations, while a tail extends up to nine generations when niches remain. Useful variants stabilize and stop mutating, so only lineages with room to specialize keep branching.

\subsection{Failure Taxonomy Details}
\label{app:taxonomy}

\begin{figure}[h]
    \centering
    \includegraphics[width=\columnwidth]{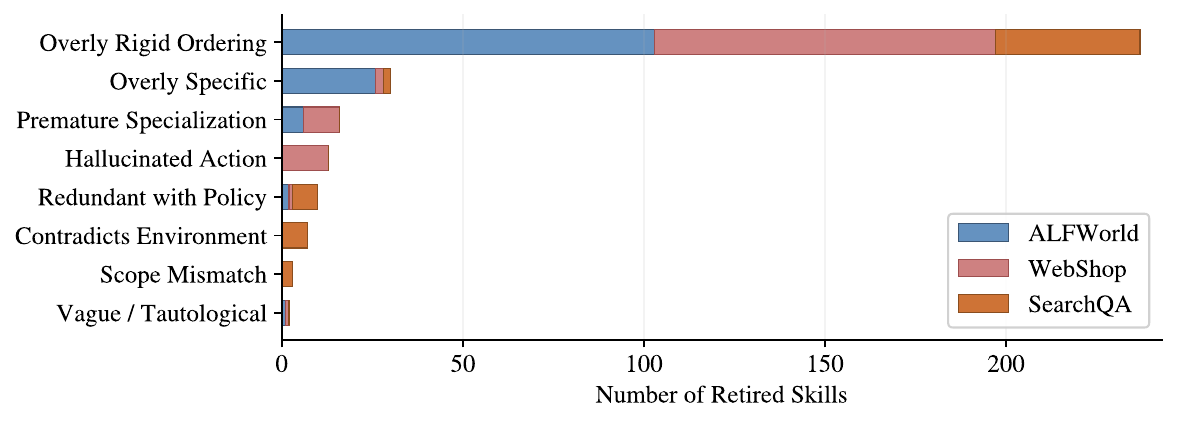}
    \caption{Failure taxonomy of 318 retirement events across three environments. Blue: ALFWorld; red: WebShop; orange: Search-Augmented QA. Overly rigid ordering dominates across all environments; environment-specific failure modes emerge in Search-Augmented QA.}
    \label{fig:taxonomy}
\end{figure}

As shown in Figure~\ref{fig:taxonomy}, the 318 retirement events were labelled into one of eight failure categories. Two annotators labelled the events under this 8-way taxonomy; an initial pass reached Cohen's $\kappa = 0.5$ inter-annotator agreement, and remaining disagreements were resolved by discussion. The dominant failure mode is \textit{overly rigid ordering} at 74.5\%, where the skill imposes a fixed procedural sequence that fails for edge cases; the second most common is \textit{overly specific} at 9.4\%, where skills work for narrow task subsets but fail broadly. Only 10 of the 318 retirement events (3.1\%) are labeled \textit{redundant with policy}; the remaining 96.9\% contain prescriptions inconsistent with successful rollouts---rigid ordering, hallucinated operations, overfit specializations, self-contradictions---so \textit{delayed obsolescence} reflects once-stable skills becoming concretely misleading, not the policy outgrowing harmless crutches. Environment-specific patterns also emerge: \textit{contradicts environment} failures appear exclusively in Search-Augmented QA, where LLM-guided mutation hallucinates search operations absent from the environment.

\subsection{WebShop: Hidden-Constraint Verification}

\begin{quote}
\small
\textbf{Parent skill} (``Confirm Hidden Specs'', gen=0): \\
\textit{Principle}: Scroll to bullet points/description to verify non-title attributes (material, care, fit, design features); reject items that satisfy title keywords but fail hidden constraints. \\
\textit{When to apply}: When evaluating candidate products against multi-attribute instructions. \\[4pt]
\textbf{Successful child} (``Systematically Verify All Hidden Constraints'', fitness=0.70, gen=1, \texttt{stable}): \\
\textit{Principle}: After selecting a candidate product, explicitly open the product description/features section and verify every constraint from the instruction that is not visible in the title or options. \\
\textit{When to apply}: When a candidate product matches title keywords but contains hidden attribute constraints. \\[4pt]
\textbf{Failed child} (``Systematically Verify Every Non-Visible Constraint'', fitness=0.21, gen=1, \texttt{retired} at step 78): \\
\textit{Principle}: After selecting a product, scroll down and click on the `Description' or `Features' tab/section to read every specification. \\
\textit{When to apply}: Whenever a product page is opened.
\end{quote}

The successful child reaches \texttt{stable} status, while the failed variant referenced interface elements such as ``click on the Description tab,'' which is absent from WebShop; its fitness decayed rapidly and the retirement gate filtered it out at step~78.

\subsection{ALFWorld: Mutation Outcomes}

The initial skill ``Pick Before You Wander'' produced two mutations with opposite outcomes:

\begin{quote}
\small
\textbf{Parent} (``Pick Before You Wander'', gen=0): \\
\textit{If the target object is visible or discovered, take it immediately; never leave it behind to explore other places.} \\[4pt]

\textbf{Successful child} (``Verify, Grab, and Confirm Pickup'', gen=1, fitness=0.70, \texttt{stable}): \\
\textit{When a needed object is visible, move directly to it, take it immediately, and then verify it is now in your inventory before proceeding.} \\
$\rightarrow$ Adds a \textbf{verification step} that catches pickup failures. \\[4pt]

\textbf{Failed child} (``Retrieve Target Before Locating Desklamp'', gen=1, fitness=0.24, \texttt{retired} at step 65): \\
\textit{First locate and pick up the target object specified in the goal before navigating to any furniture.} \\
$\rightarrow$ Introduces a rigid ordering that fails for tasks where the lamp must be found first.
\end{quote}

From the same parent, mutation produced both a beneficial specialization (adding a verification step) and a harmful over-specification (imposing a rigid task ordering). The retirement mechanism correctly identified and removed the harmful variant.

\subsection{Search-Augmented QA: Query Reformulation Lineage}

A multi-generation forging chain on Search-Augmented QA:

\begin{quote}
\small
\textbf{Gen-0}: ``Disambiguation Add-Ons'' \\
\textit{If the name is ambiguous, append clarifiers (birth year, nationality, known work) to the query.} \\[3pt]
$\downarrow$ mutation \\[3pt]
\textbf{Gen-1}: ``Structured Query Escalation with Fallback'' (fitness=0.21, \texttt{retired} at step 83) \\
\textit{Systematically escalate through three reformulation strategies when initial results lack evidence.} \\
$\rightarrow$ Too rigid; forced 3-step escalation even for simple queries. \\[3pt]
$\downarrow$ later mutation from same lineage \\[3pt]
\textbf{Gen-1}: ``Verify Before Acting'' (fitness=0.70, \texttt{stable}) \\
\textit{Before performing any action on an entity, first confirm its identity by searching with disambiguating context.} \\
$\rightarrow$ Simpler, more general, and more effective.
\end{quote}

Forging explores multiple directions from the same parent: the overly complex variant is retired while the simpler, more robust variant survives.

\section{KL Regularization under a Library-Conditioned Context}
\label{app:kl_note}

This appendix elaborates on how the KL penalty in Eq.~\eqref{eq:grpo} composes with the evolving skill library, complementing the non-stationarity of the fitness estimator formalized in Eq.~\eqref{eq:nonstationary}.

Under our formulation (Eq.~\eqref{eq:context}), the library $\mathcal{S}^{(k)}$ enters the policy through the context $p_t$ rather than through the policy parameters $\theta$. The parameterization of $\pi_\theta$ is therefore invariant across forging cycles; what varies over training is the prompt distribution fed to the fixed network. The KL term $D_{\text{KL}}(\pi_\theta \| \pi_\text{ref})$ is always evaluated at the same context $c_{n,l}$ seen by both $\pi_\theta$ and $\pi_\text{ref}$, so the divergence is well-defined at every training step, regardless of how far $\mathcal{S}^{(k)}$ has drifted from $\mathcal{S}^{(0)}$.

Two structural properties align the fixed anchor $\pi_\text{ref}$ with the library-conditioned inputs it encounters during RL. First, mutated skills are teacher-model rewrites of surviving parents that preserve the $(v, b, \omega)$ schema of Section~\ref{sec:skill_context}, so new prompts differ from SFT-time prompts mainly in skill content rather than in structural form, on which $\pi_\text{ref}$ remains a calibrated next-token predictor. Second, we use $\beta = 10^{-3}$ (Table~\ref{tab:training}), so the KL term acts as a soft regularizer against large single-step deviations rather than a tight anchor to the SFT policy's full conditional distribution. An alternative design is to re-anchor $\pi_\text{ref}$ at each forging cycle, trading the clean separation between SFT initialization and RL optimization for tighter alignment between the reference policy and the current library; this variant is compatible with the rest of our framework, and the fixed-anchor choice is the simpler default we adopt here.

\section{Discussion}
\label{app:discussion}

\paragraph{Design choices.}
Fitness is tracked as a cumulative running average (Eq.~\eqref{eq:fitness}), which integrates evidence over the full lifecycle of each skill. This is the natural estimator for our detection target, skills whose accumulated success record needs to be distinguished from skills that were never effective; it is what makes ``once stable, now failing'' a well-defined pattern to catch. A faster estimator such as an exponential moving average would respond more quickly to fitness changes, but this sensitivity cuts both ways: a skill that encounters a transiently hard batch of tasks would be flagged before the policy has stabilized, producing false retirements; empirically, the moving average lowers average success by 1.0 point (Table~\ref{tab:design_alternatives}). The cumulative average's inertia is intentional---the lifecycle's usage-count gates ($N_{\text{retire}}$, $N_{\text{stable}}$) already guard against premature decisions; within that framework, the cumulative estimate reliably separates skills with a strong historical record from those that have consistently underperformed. Table~\ref{tab:skillfurnace} reports an average retirement-event lifespan of 38.0 training steps, showing that retirement occurs promptly relative to the 200-step training budget despite the conservative estimator.

\paragraph{Mutation survival.}
Across the three environments, 213 mutant skills were generated: 88 in ALFWorld, 55 in WebShop, and 70 in Search-Augmented QA. The majority survive the lifecycle: the mutant survival rate is 64.8\%, 52.7\%, and 62.9\% respectively, so LLM-guided mutation produces viable skills more often than not. Each forging cycle requires up to $K_{\text{mutate}}$ teacher model calls for mutation plus lightweight fitness bookkeeping.

\paragraph{Ablation library dynamics.}
In Table~\ref{tab:ablation} the $-$mutation and $-$forging variants both retain the initial library size (44 on ALFWorld, 66 on WebShop, 41 on Search-Augmented QA). These numbers refer to $\lvert\mathcal{S}_0\rvert$, the distilled library \emph{before} pre-retirement, to make cross-variant comparison direct. For $-$forging the entire forging method, including pre-retirement, runtime retirement, and mutation, is disabled, so the SFT data is unfiltered and no skills are added or removed after initialization. For $-$mutation, pre-retirement and runtime retirement both run, but the two mechanisms remove very few skills on these environments in practice: pre-retirement at $\delta_{\text{pre}}=0.3$ with $N_{\text{min}}=3$ discards only initial skills that demonstrably hurt the base-model policy, and on the three environments this is a small fraction of $\mathcal{S}_0$ (typically 0--3 skills); the remaining skills then accumulate enough usage during RL to stay above $\delta_{\text{retire}}$, so runtime retirement triggers rarely when mutation is off and no low-fitness children are being introduced. Size changes driven by the lifecycle therefore manifest primarily once mutation is re-enabled, which is consistent with the joint role of mutation and retirement reported in Section~\ref{sec:ablation}. The $-$pre-retirement variant confirms this picture from the opposite direction: with runtime retirement and mutation intact, the final library still settles at the cap (100, 95, 85) because the cap is the binding constraint once mutation is active. The 1.2--3.5\% success drop therefore arises not from library size but from SFT data quality: pre-retirement is an integral first stage of the lifecycle that shapes the initial policy $\pi_{\theta_\text{sft}}$ on which the subsequent retirement and mutation operate.

The $-$retirement column in Table~\ref{tab:ablation} reaches 132, 121, and 111 skills on the three environments, matching the \emph{total-ever-created} count of the full method. This follows from the forging schedule: the number of forging cycles and the per-cycle budget $K_{\text{mutate}}$ are fixed, so the cumulative number of child skills generated is largely invariant to whether low-fitness parents are subsequently removed. Retirement determines which skills remain in the active library at any point, not how many are produced during training; correspondingly the full method's active library settles at a much smaller size (100, 95, 85) while $-$retirement accumulates every generated skill.

\paragraph{Evaluation protocol.}
All reported numbers use stochastic decoding at temperature $1.0$ for both training-time rollouts and test-time evaluation, so that evaluation noise reflects the policy's deployed behavior rather than a degenerate argmax. Each reported score is averaged over five evaluation runs of the best validation checkpoint with sampling seeds $\{30, 40, 42, 50, 62\}$. The mean $\pm$ standard deviation of the overall success rate is $92.4 \pm 0.5\%$ on ALFWorld, $78.4 \pm 0.4\%$ on WebShop, and $48.7 \pm 0.3\%$ on Search-Augmented QA, with the low variance confirming that reported results are stable under resampling.

\paragraph{Statistical significance.}
We test whether \method's overall success rate exceeds the strongest memory-augmented RL baseline, SkillRL, using a one-sided one-sample $t$-test against the SkillRL point estimate under the null $H_0\!:\mu_{\method}\!=\!\mu_{\text{SkillRL}}$ with our five-seed evaluation sample ($n\!=\!5$, $\mathrm{df}\!=\!4$). On ALFWorld ($t\!=\!11.18$, $p\!=\!1.8\!\times\!10^{-4}$), WebShop ($t\!=\!31.86$, $p\!=\!2.9\!\times\!10^{-6}$), and Search-Augmented QA ($t\!=\!14.16$, $p\!=\!7.2\!\times\!10^{-5}$), the improvement is significant at $p<10^{-3}$ on all three environments. The corresponding two-sided $95\%$ confidence intervals, $[91.78, 93.02]$, $[77.90, 78.90]$, and $[48.33, 49.07]$, do not overlap the SkillRL baseline values of $89.9$, $72.7$, and $46.8$, so the $1.9$--$5.7\%$ absolute gains are unlikely to be explained by sampling noise. Under the conservative assumption that SkillRL's per-seed variance matches our observed per-environment standard deviations of $0.5\%$, $0.4\%$, and $0.3\%$ respectively, a pooled Welch's $t$-test yields $t=7.91$ (ALFWorld), $t=22.51$ (WebShop), and $t=10.03$ (Search-Augmented QA), with $p<10^{-4}$ on all three environments, preserving the conclusion.

\paragraph{Broader impact.}
Treating experience accumulation as a quality-control problem improves agent performance and yields interpretable artifacts: the \skillfurnace dataset enables follow-up research on skill quality prediction and failure-aware generation. The contributed collection contains only SFT trajectories, evolved skill libraries with fitness annotations, and retirement events with failure annotations, and introduces no risks beyond those inherent to LLM agents that act in interactive environments.

\section{Additional Controlled Experiments}
\label{app:extra}

This appendix reports controlled experiments that complement the ablation of Section~\ref{sec:ablation}. Unless stated otherwise, entries are success rate on ALFWorld and WebShop and accuracy on Search-Augmented QA (\%), each the mean $\pm$ standard deviation over the five evaluation sampling seeds at the best validation checkpoint, following the protocol of Appendix~\ref{app:discussion}. \textbf{Avg} is the unweighted mean over the three environments. In every experiment all settings other than the one under study are held at their defaults, and differences smaller than the per-seed deviations are not read as an ordering.

\paragraph{Matched teacher budget.}
Among the lifecycle mechanisms only mutation uses the external teacher. We count the teacher rewrites $B$ that \method performs over training, 88, 55, and 70 on ALFWorld, WebShop, and Search-Augmented QA, equal to the number of mutation children, and grant SkillRL the same $B$ rewrites appended to its library with no lifecycle gate, that is, no \texttt{trial} admission and no fitness-driven retirement. Because SkillRL maintains no fitness scores, the $B$ parents are sampled in proportion to retrieval frequency under SkillRL's own rollouts, so the control uses neither \method's fitness nor its lifecycle states. The gap between \method and the teacher-matched SkillRL measures the lifecycle at equal teacher budget, and the gap between the teacher-matched and the plain SkillRL measures distillation alone (Table~\ref{tab:teacher_budget}). Averaged over the three environments, teacher distillation alone accounts for about 0.7 points and the lifecycle for about 2.7 points at a matched teacher budget. The lifecycle also keeps \method's library at the cap $S_{\text{max}}$, whereas the teacher-matched SkillRL retains every skill it receives and ends with a larger, unpruned library. Mutation's small additive contribution in Table~\ref{tab:ablation}, 0.5 to 1.6 points, understates its role: retirement removes low-fitness skills, while mutation replenishes the library with improved variants so that pruning does not merely shrink coverage.

\begin{table}[h]
\caption{Matched teacher budget. Rewrites: number of teacher rewrites; $B$ is the number performed by \method in each environment.}
\label{tab:teacher_budget}
\centering
\small
\setlength{\aboverulesep}{2pt}
\setlength{\belowrulesep}{2pt}
\setlength{\tabcolsep}{4pt}
\begin{tabular}{lcccccc}
\toprule
\textbf{Method} & \textbf{Rewrites} & \textbf{Lifecycle} & \textbf{ALFWorld} & \textbf{WebShop} & \textbf{Search-QA} & \textbf{Avg} \\
\midrule
SkillRL & 0 & no & $89.9{\pm}0.5$ & $72.7{\pm}0.4$ & $46.8{\pm}0.3$ & $69.8{\pm}0.2$ \\
SkillRL + teacher rewrites & $B$ & no & $90.3{\pm}0.5$ & $74.2{\pm}0.5$ & $47.0{\pm}0.3$ & $70.5{\pm}0.3$ \\
\ours \method & \ours $B$ & \ours yes & \ours $\best{92.4}{\pm}0.5$ & \ours $\best{78.4}{\pm}0.4$ & \ours $\best{48.7}{\pm}0.3$ & \ours $\best{73.2}{\pm}0.2$ \\
\bottomrule
\end{tabular}
\end{table}

\paragraph{Teacher capability.}
We vary only the mutation teacher $\mathcal{M}_T$. Besides the default Kimi-K2.5 we test two other capable instruction models, GLM-5.2~\citep{glm52} and MiniMax M3~\citep{minimaxm3}, and, as the strongest control, self-mutation, in which the base model itself serves as the teacher so that no external knowledge is injected. Disabling mutation while keeping retirement gives the floor (Table~\ref{tab:teacher_capability}). Capable external teachers stay within 0.3 points of the default, while self-mutation falls back near the retirement-only floor and still exceeds SkillRL's 69.8 in Table~\ref{tab:teacher_budget}.

\begin{table}[h]
\caption{Teacher capability for LLM-guided mutation.}
\label{tab:teacher_capability}
\centering
\small
\setlength{\aboverulesep}{2pt}
\setlength{\belowrulesep}{2pt}
\setlength{\tabcolsep}{9pt}
\begin{tabular}{lcccc}
\toprule
\textbf{Mutation teacher $\mathcal{M}_T$} & \textbf{ALFWorld} & \textbf{WebShop} & \textbf{Search-QA} & \textbf{Avg} \\
\midrule
None (retirement only) & $91.8{\pm}0.5$ & $76.8{\pm}0.4$ & $48.2{\pm}0.3$ & $72.3{\pm}0.2$ \\
Base model (self-mutation) & $91.5{\pm}0.5$ & $77.1{\pm}0.5$ & $48.0{\pm}0.4$ & $72.2{\pm}0.3$ \\
GLM-5.2 & $92.3{\pm}0.4$ & $77.9{\pm}0.4$ & $48.5{\pm}0.3$ & $72.9{\pm}0.2$ \\
MiniMax M3 & $92.5{\pm}0.5$ & $78.1{\pm}0.4$ & $48.9{\pm}0.3$ & $73.2{\pm}0.2$ \\
\ours Kimi-K2.5 (default) & \ours $92.4{\pm}0.5$ & \ours $78.4{\pm}0.4$ & \ours $48.7{\pm}0.3$ & \ours $73.2{\pm}0.2$ \\
\bottomrule
\end{tabular}
\end{table}

\paragraph{Credit assignment.}
Eq.~\eqref{eq:credit} attributes one binary reward to all $K$ co-retrieved skills. We test this approximation in two ways (Table~\ref{tab:credit}). The $K=1$ stress test retrieves a single skill per episode, so each reward is attributed exactly; it changes retrieval breadth and credit assignment together, so its success-rate gap is not by itself a clean measure of credit error. The per-skill attribution variant keeps the default $K=6$, so the agent's context is unchanged, and replaces Eq.~\eqref{eq:credit} with each skill's leave-one-out marginal contribution over its co-retrieval sets, which then drives all lifecycle transitions in place of shared fitness. For both we report the Spearman correlation of the resulting per-skill fitness ranking with the default and the agreement of retirement decisions; for per-skill attribution we also report the agreement of mutation-candidate selection.

\begin{table}[h]
\caption{Credit assignment. $\rho$: Spearman correlation of the per-skill fitness ranking with the default shared reward of Eq.~\eqref{eq:credit}; Ret./Mut.: agreement of retirement and mutation-candidate decisions with the default. ALF./Web./QA: ALFWorld, WebShop, Search-Augmented QA.}
\label{tab:credit}
\centering
\small
\setlength{\aboverulesep}{2pt}
\setlength{\belowrulesep}{2pt}
\setlength{\tabcolsep}{3pt}
\begin{tabular}{lccccccccc}
\toprule
\textbf{Estimator} & $K$ & \textbf{ALF.} & \textbf{Web.} & \textbf{QA} & \textbf{Avg} & $\rho$ & \textbf{Ret.} & \textbf{Mut.} \\
\midrule
\ours Shared (default) & \ours 6 & \ours $92.4{\pm}0.5$ & \ours $78.4{\pm}0.4$ & \ours $48.7{\pm}0.3$ & \ours $73.2{\pm}0.2$ & \ours -- & \ours -- & \ours -- \\
Single skill & 1 & $91.8{\pm}0.5$ & $76.9{\pm}0.5$ & $48.4{\pm}0.3$ & $72.4{\pm}0.3$ & 0.86 & 89\% & -- \\
Leave-one-out & 6 & $92.7{\pm}0.5$ & $77.9{\pm}0.5$ & $48.8{\pm}0.4$ & $73.1{\pm}0.3$ & 0.91 & 93\% & 88\% \\
\bottomrule
\end{tabular}
\end{table}

Pooled over the three environments, per-skill attribution and the default reach the same retirement decision on 269 skills and different decisions on 21. The 21 disagreeing skills lie close to the threshold, with mean $|f_i - \delta_{\text{retire}}|$ of 0.07 against 0.31 for the agreeing ones, and are rarely used, with mean usage $U_i$ of 23 against 48. Reassigning only these decisions within the default run, a targeted counterfactual rather than end-to-end retraining, changes aggregate success by $+0.1$. The shared-reward approximation therefore does not materially affect the lifecycle.

\paragraph{Design alternatives.}
Table~\ref{tab:design_alternatives} compares the main alternatives for three load-bearing design choices, whose rationale is given in Appendices~\ref{app:hyperparams} and~\ref{app:discussion}: the fitness estimator, the granularity of the lifecycle, and the weight used to sample mutation parents. Each alternative lowers average success relative to the default, most clearly the two-state lifecycle and the EMA, at 1.7 and 1.0 points of average success.

\begin{table}[h]
\caption{Alternatives for three design choices of \method. EMA: exponential moving average; two states: \texttt{active} and \texttt{retired} only.}
\label{tab:design_alternatives}
\centering
\small
\setlength{\aboverulesep}{2pt}
\setlength{\belowrulesep}{2pt}
\setlength{\tabcolsep}{4pt}
\begin{tabular}{llcccc}
\toprule
\textbf{Design choice} & \textbf{Variant} & \textbf{ALFWorld} & \textbf{WebShop} & \textbf{Search-QA} & \textbf{Avg} \\
\midrule
\multirow{2}{*}{Fitness} & \ours cumulative (default) & \ours $92.4{\pm}0.5$ & \ours $78.4{\pm}0.4$ & \ours $48.7{\pm}0.3$ & \ours $73.2{\pm}0.2$ \\
 & EMA & $91.1{\pm}0.5$ & $77.2{\pm}0.5$ & $48.4{\pm}0.3$ & $72.2{\pm}0.3$ \\
\midrule
\multirow{2}{*}{Lifecycle} & \ours four states (default) & \ours $92.4{\pm}0.5$ & \ours $78.4{\pm}0.4$ & \ours $48.7{\pm}0.3$ & \ours $73.2{\pm}0.2$ \\
 & two states & $90.8{\pm}0.5$ & $76.1{\pm}0.5$ & $47.7{\pm}0.4$ & $71.5{\pm}0.3$ \\
\midrule
\multirow{3}{*}{Mut. weight} & \ours $1-f_i$ (default) & \ours $92.4{\pm}0.5$ & \ours $78.4{\pm}0.4$ & \ours $48.7{\pm}0.3$ & \ours $73.2{\pm}0.2$ \\
 & $1/f_i$ & $91.9{\pm}0.5$ & $77.3{\pm}0.5$ & $48.5{\pm}0.4$ & $72.6{\pm}0.3$ \\
 & uniform & $92.5{\pm}0.5$ & $77.5{\pm}0.4$ & $48.0{\pm}0.4$ & $72.7{\pm}0.3$ \\
\bottomrule
\end{tabular}
\end{table}

\ifcameraReady
\input{checklist}
\fi

\end{document}